%% file: main.tex
\documentclass[10pt,twocolumn,letterpaper]{article}

\usepackage{cvpr}              

\input{preamble}

\definecolor{cvprblue}{rgb}{0.21,0.49,0.74}
\usepackage[pagebackref,breaklinks,colorlinks,citecolor=cvprblue]{hyperref}
\usepackage[accsupp]{axessibility} 

\def\paperID{375} 
\def\confName{CVPR}
\def\confYear{2024}

\title{MultiPly: Reconstruction of Multiple People from Monocular Video in the Wild}

\author{Zeren Jiang$^{*1}$ \quad Chen Guo$^{*1}$ \quad Manuel Kaufmann$^{1}$ \quad Tianjian Jiang$^{1}$ \quad \\ Julien Valentin$^{2}$ \quad Otmar Hilliges$^{1}$ \quad Jie Song$^{1}$ \\
 $^1$ETH Z{\"u}rich \quad 
 $^2$Microsoft \\
}

\begin{document}
\twocolumn[{%
\renewcommand\twocolumn[1][]{#1}%
\maketitle

\vspace{-3.5em}
\begin{center}
    \captionsetup{type=figure}
   \includegraphics[width=\linewidth,trim=0 5 0 0,clip]{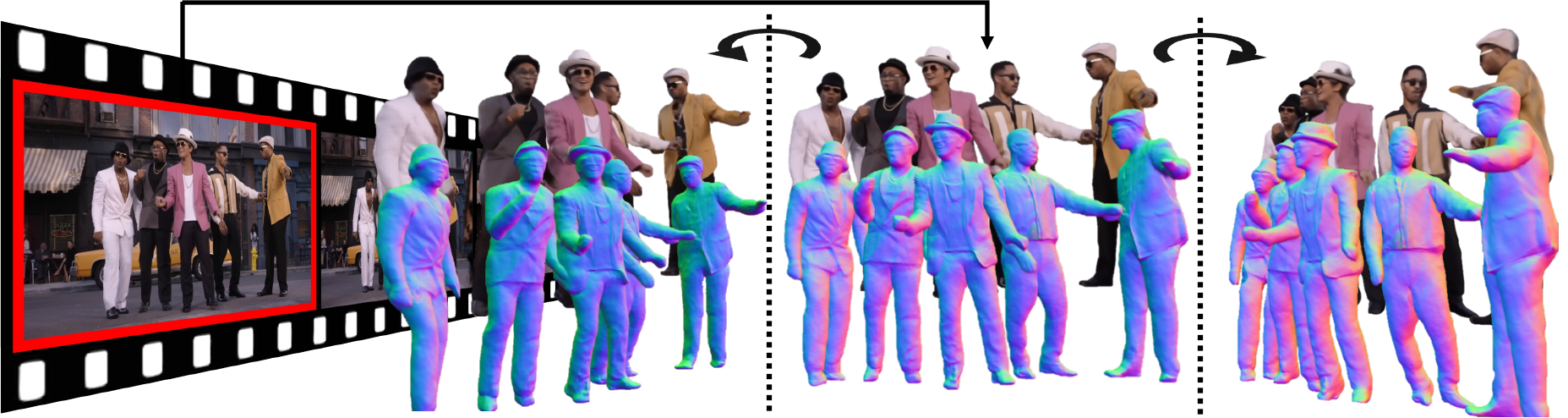}
    \captionof{figure}{We propose \methodname, a novel framework to reconstruct multiple people in 3D from in-the-wild monocular videos. Our method can recover the complete 3D human with high-fidelity shape and appearance, even in scenarios involving occlusions and close interactions.}
    \label{fig:teaser}

\end{center}%

}]

\def\thefootnote{*}\footnotetext{These authors contributed equally to this work}

\input{figures.tex}
\input{tables.tex}
\input{alg}

\input{sec/00_abstract}    
\input{sec/01_introduction}
\input{sec/02_related_work}
\input{sec/03_method}

\input{sec/04_experiment}

\input{sec/05_conclusion}
\newpage
{
    \small
    \bibliographystyle{ieeenat_fullname}
    \bibliography{main}
}

\end{document}

%% file: preamble.tex
\usepackage[dvipsnames]{xcolor}

\newcommand{\todo}[1]{{\color{red}#1}}

\usepackage{graphicx}
\usepackage{amsmath}
\usepackage{amssymb}
\usepackage{booktabs}
\usepackage{url}
\usepackage{enumitem}
\usepackage{xspace}
\usepackage{paralist}
\usepackage{multirow}
\usepackage{adjustbox}
\usepackage{multibib}
\makeatletter

\newcommand{\Rmnum}[1]{\expandafter\@slowromancap\romannumeral #1@}
\makeatother

\usepackage{bm}

\newif\ifshowcomments
\showcommentstrue %

\ifshowcomments
    \newcommand{\otmar}[1]{\noindent\textcolor{ForestGreen}{\textbf{[Otmar:~}#1\textbf{]}}}
    \newcommand{\jie}[1]{\noindent\textcolor{Dandelion}{\textbf{[Jie:~}#1\textbf{]}}}
    \newcommand{\MK}[1]{\noindent\textcolor{teal}{\textbf{[Manuel:~}#1\textbf{]}}}
    \newcommand{\OH}[1]{{\color{blue}[OH: #1]}}
    \newcommand{\pmnote}[1]{\PM{#1}}
    \newcommand{\oh}[1]{\OH{#1}}
    \newcommand{\cg}[1]{{\color{magenta}[CG: #1]}}
    \newcommand{\JV}[1]{{\color{red}[JV: #1]}}
    \newcommand{\zr}[1]{{\color{ForestGreen}[ZR: #1]}}
\else
    \newcommand{\todo}[1]{\unskip}
    \newcommand{\otmar}[1]{\unskip}
    \newcommand{\jie}[1]{\unskip}
    \newcommand{\MK}[1]{\unskip}
    \newcommand{\OH}[1]{\unskip}
    \newcommand{\pmnote}[1]{\unskip}
    \newcommand{\oh}[1]{\unskip}
    \newcommand{\cg}[1]{\unskip}
    \newcommand{\JV}[1]{\unskip}
    \newcommand{\zr}[1]{\unskip}

\fi

\newcommand{\Ra}[1]{{\textbf{\color{magenta}{R1}}}}
\newcommand{\Rb}[1]{{\textbf{\color{blue}{R2}}}}
\newcommand{\Rc}[1]{{\textbf{\color{cyan}{R3}}}}
\newcommand{\methodname}{MultiPly\xspace}
\newcommand{\suppmat}{Supp. Mat\xspace}

\newcommand{\figref}[1]{Fig.~\ref{#1}}
\newcommand{\tabref}[1]{Tab.~\ref{#1}}
\newcommand{\secref}[1]{Sec.~\ref{#1}}

\definecolor{babyblue}{rgb}{0.54, 0.81, 0.94}

%% file: figures.tex
\newcommand{\figurePipeline}{

\begin{figure*}[t]
    \centering
    \includegraphics[width=\textwidth]{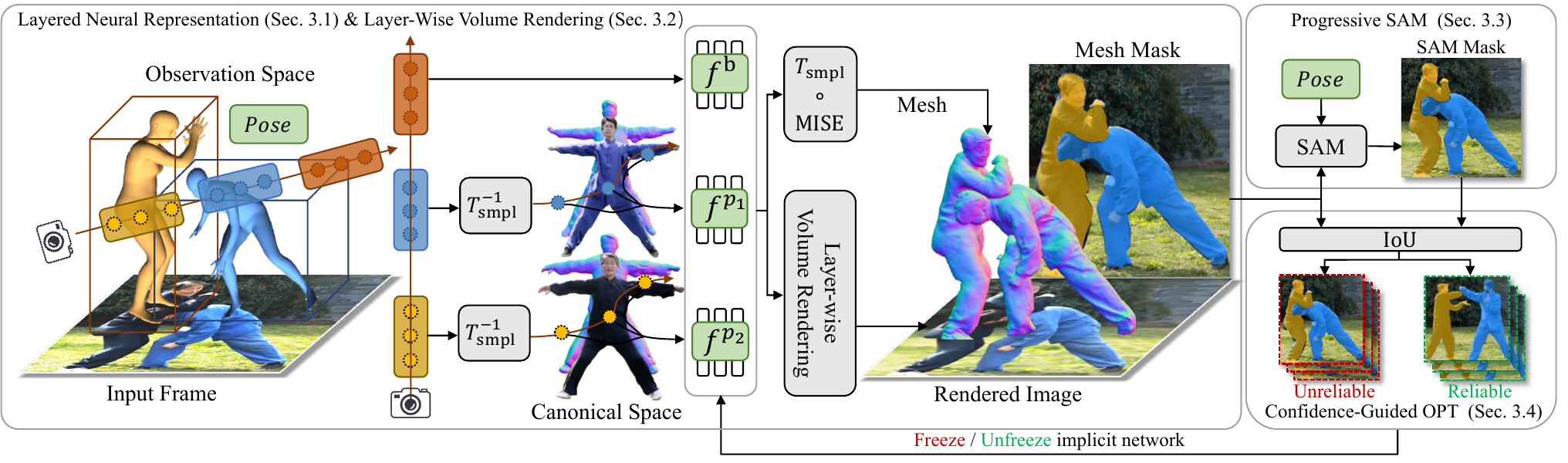}
    \vspace{-0.5cm}
    \caption{\textbf{Method overview.} Given an image and SMPL estimation, we sample human points along the camera ray based on the bounding boxes of SMPL bodies and the background points based on NeRF++. We warp sampled human points into canonical space via inverse warping and evaluate the person-specific implicit network to obtain the SDF and radiance values (\secref{sec:representation}). The layer-wise volume rendering is then applied to learn the implicit networks from images (\secref{sec:volume}). We build a closed-loop refinement for instance segmentation by dynamically updating prompts for SAM using evolving human models (\secref{sec:sam_prompt}). Finally, we formulate a confidence-guided optimization that only optimizes pose parameters for unreliable frames and jointly optimizes pose \emph{and} implicit networks for reliable frames (\secref{sec:delay}).}
    \label{fig:method}
\vspace{-0.2cm}
\end{figure*}
}

\newcommand{\figureReconComp}{

\begin{figure*} 
    \centering
    \includegraphics[width=\textwidth,trim=0 7 0 0,clip]{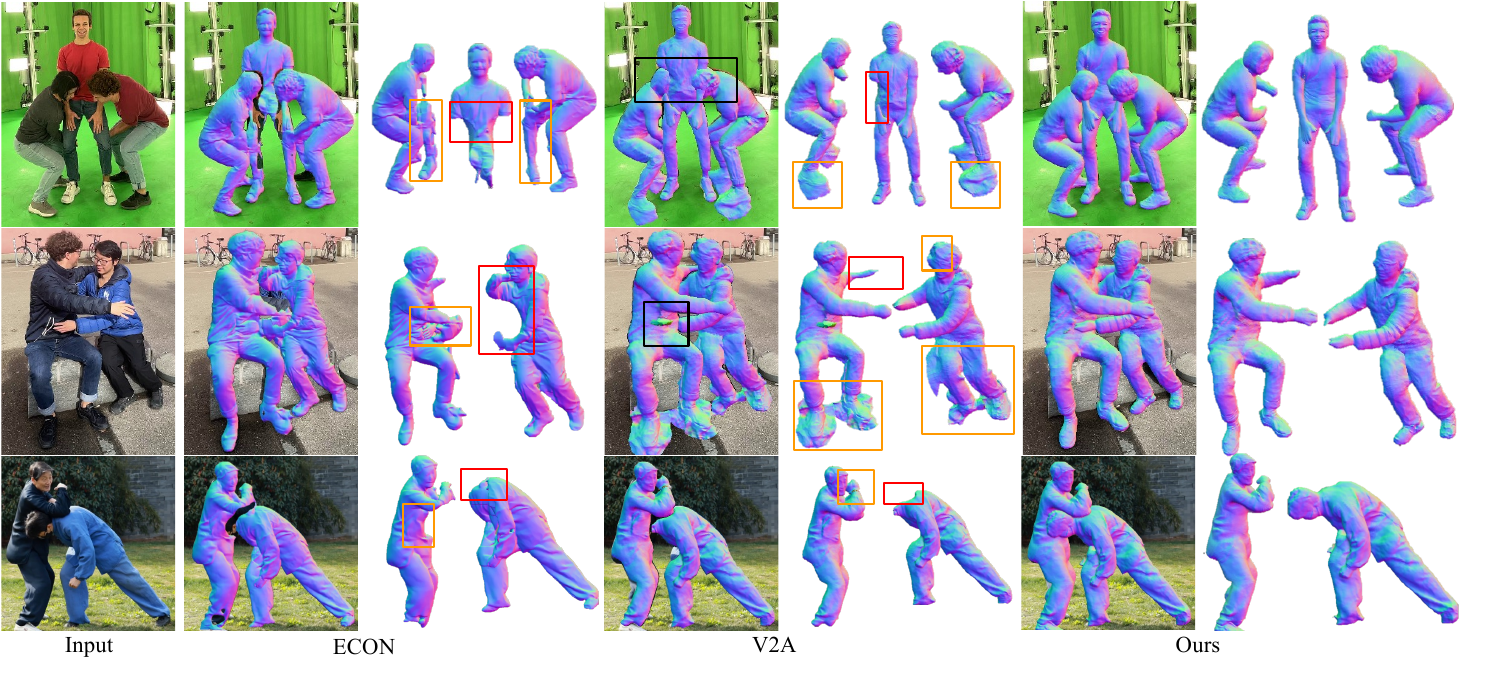}
    \vspace{-0.5cm}
    \caption{\textbf{Qualitative reconstruction comparison.} We show both the overlaid and separated reconstruction results for each method. \textbf{\textcolor[RGB]{255,0,0}{Red}} bounding boxes: the incomplete reconstruction of the occluded part. \textbf{\textcolor[RGB]{255,125,0}{Orange}} bounding boxes: incorrect instance segmentation results caused by the surrounding visual complexities. \textbf{Black} bounding boxes: inaccurate spatial arrangement due to pose error.
    }
    \label{fig:recon}

\end{figure*}

}

\newcommand{\figureSupReconComp}{

\begin{figure*}[t]
    \centering
    \includegraphics[width=\textwidth,trim=0 7 0 0,clip]{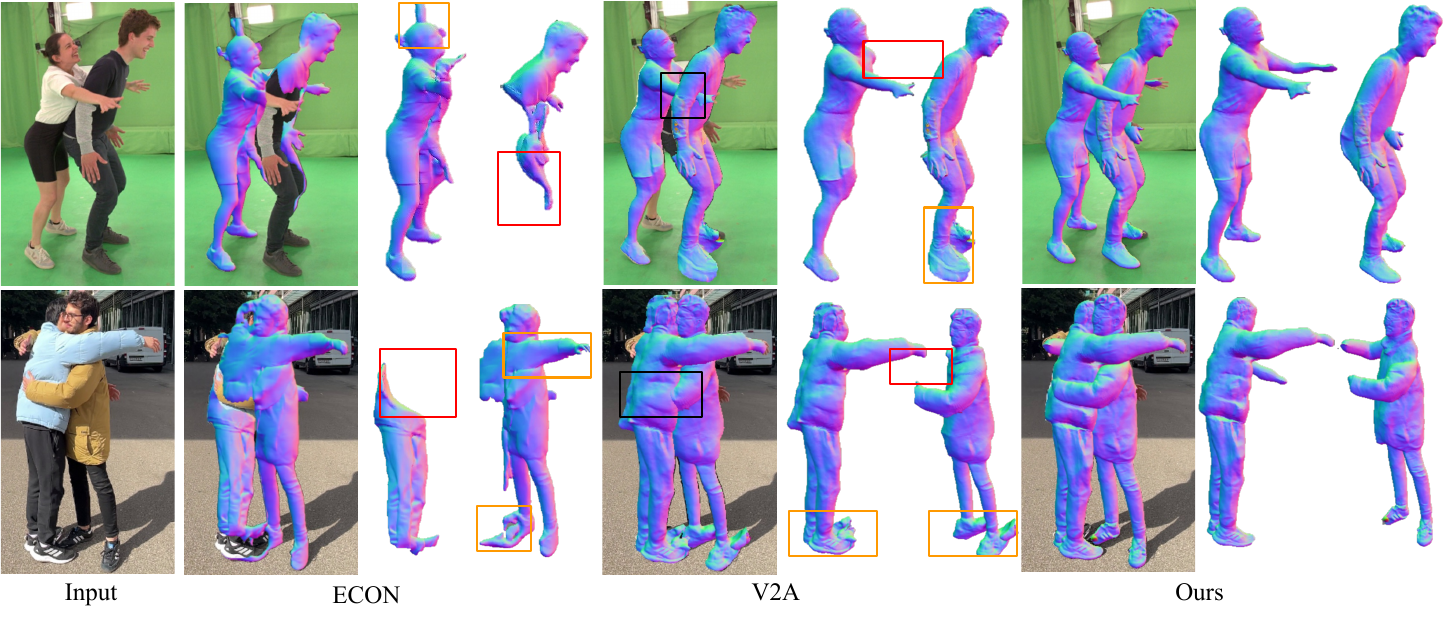}
    \caption{\textbf{Additional qualitative reconstruction comparison.} We show both the overlaid and separated reconstruction results for each method. \textbf{\textcolor[RGB]{255,0,0}{Red}} bounding boxes: the incomplete reconstruction of the occluded part. \textbf{\textcolor[RGB]{255,125,0}{Orange}} bounding boxes: incorrect instance segmentation results caused by the surrounding visual complexities. \textbf{Black} bounding boxes: inaccurate spatial arrangement due to pose error.
    }
    \label{fig:sup_recon}

\end{figure*}

}

\newcommand{\figureAblation}{

\begin{figure}[t]
    
    \begin{subfigure}{60pt}
        \centering
        \includegraphics[width=60pt, height=98.8pt]{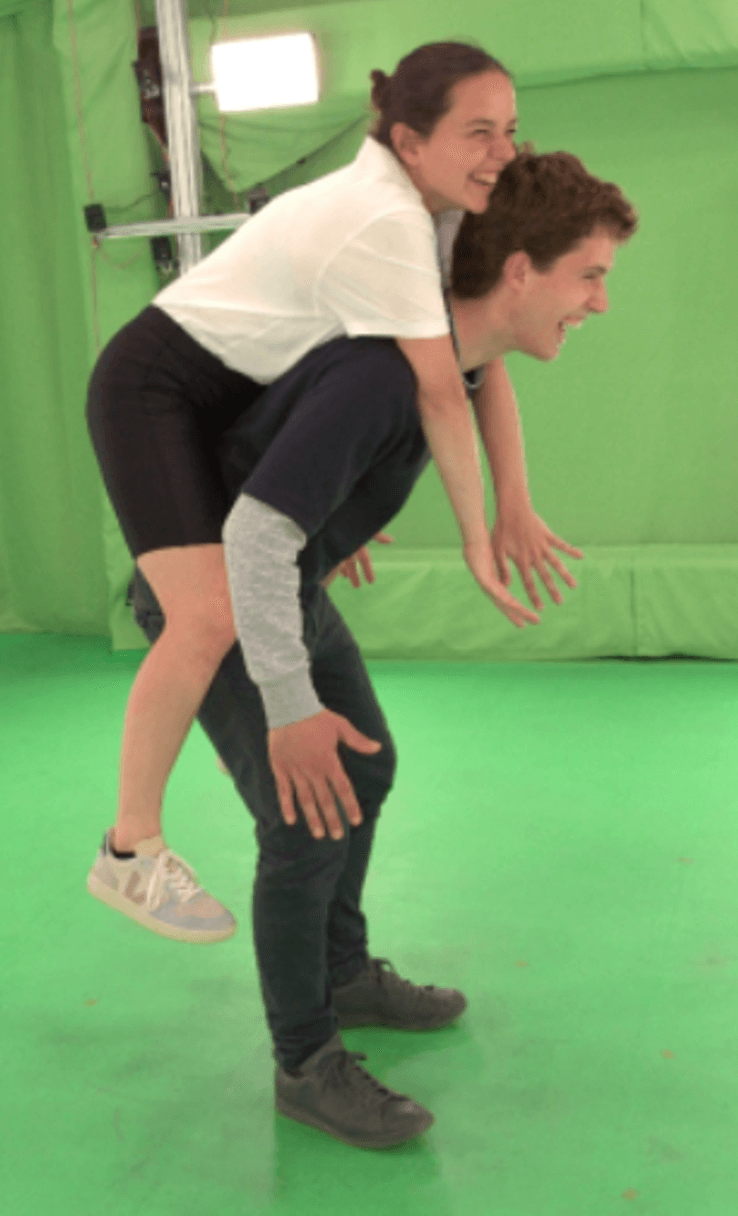}
    \end{subfigure}
    \hspace{-0.2cm}
    \begin{subfigure}{60pt}
        \centering
        \includegraphics[width=60pt, height=98.8pt]{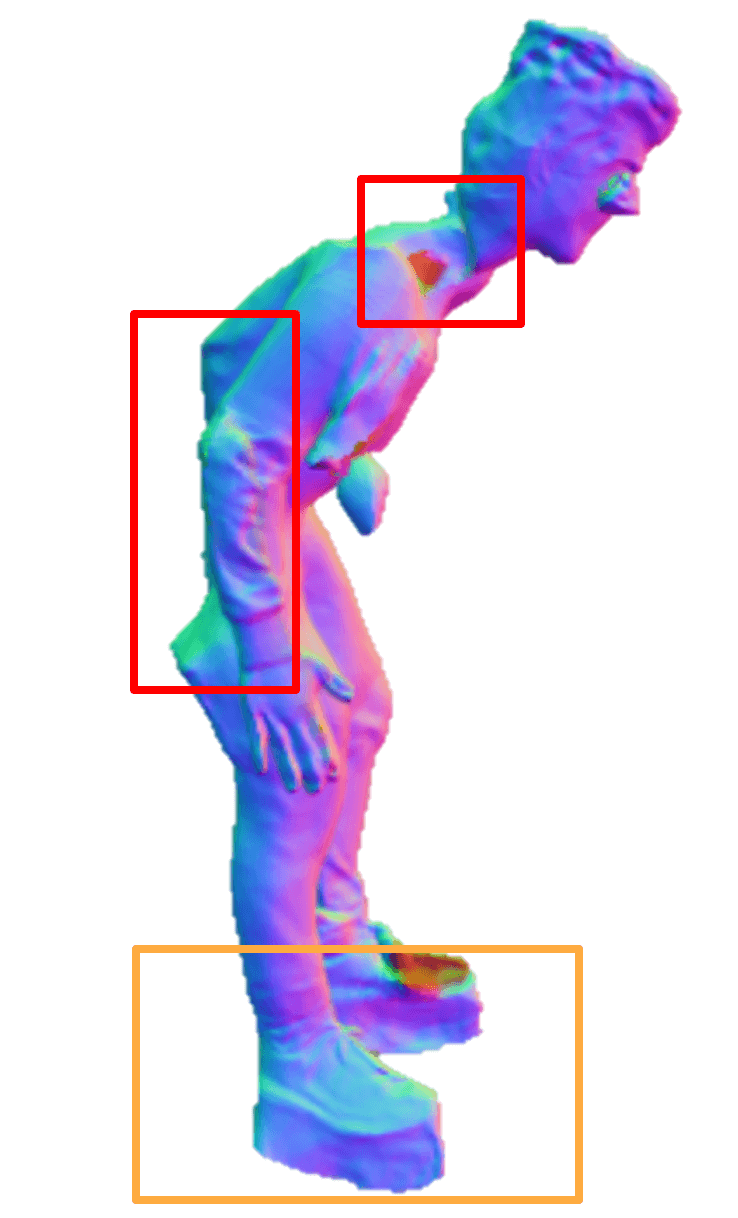}
    \end{subfigure}
    \hspace{-0.2cm}
    \begin{subfigure}{60pt}
        \centering
        \includegraphics[width=60pt, height=98.8pt]{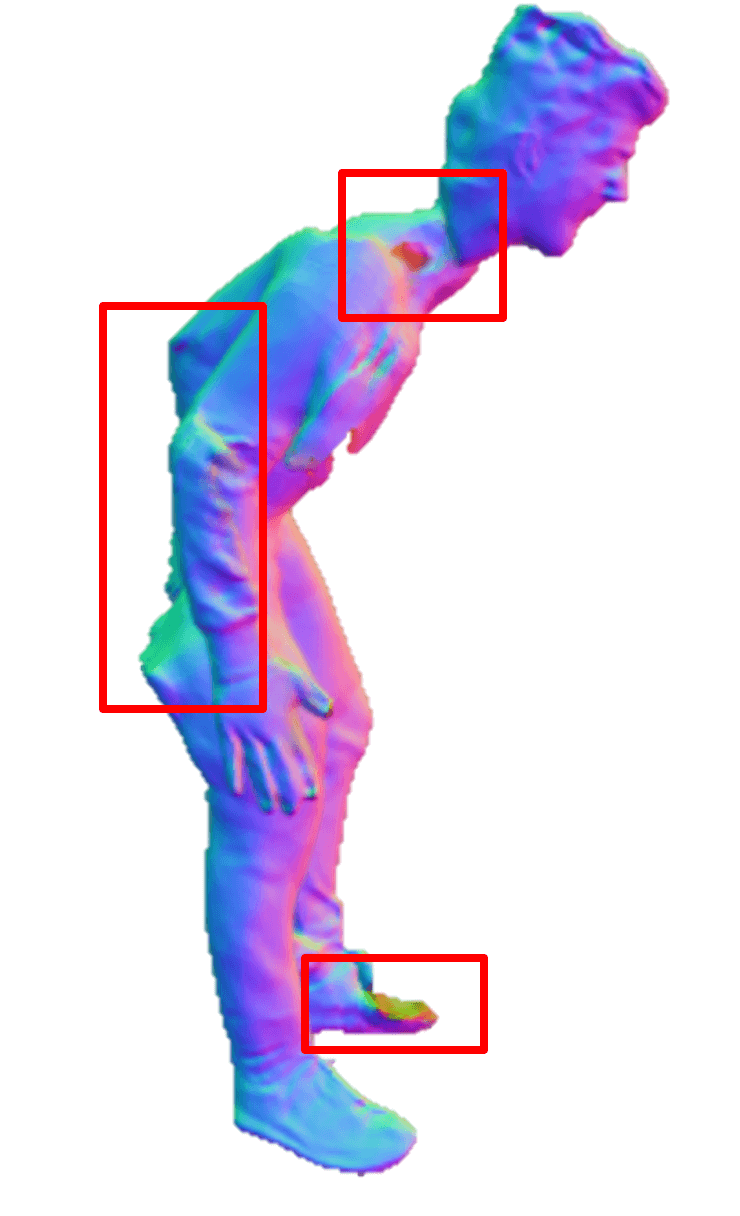}
    \end{subfigure}
    \hspace{-0.2cm}
    \begin{subfigure}{60pt}
        \centering
        \includegraphics[width=60pt, height=98.8pt]{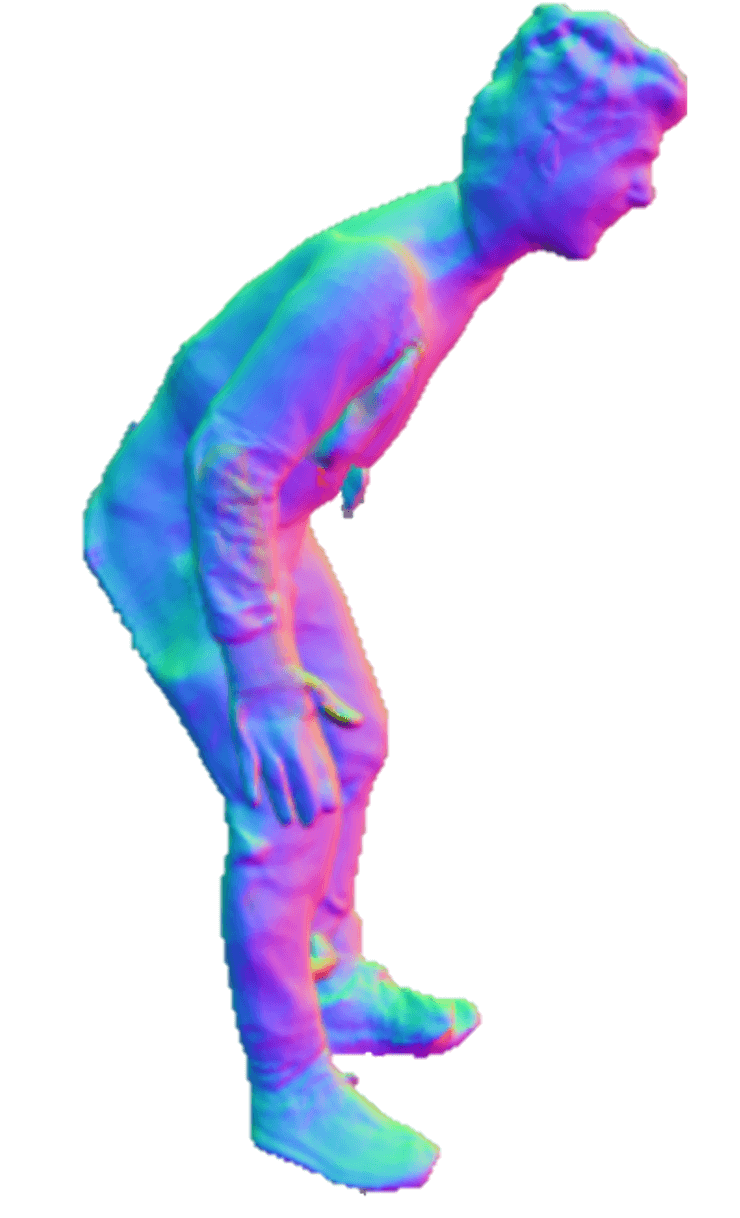}
    \end{subfigure}
    \vspace{-0.47cm}
    \\    
    \begin{subfigure}{60pt}
        \centering
        \includegraphics[width=60pt, height=98.8pt]{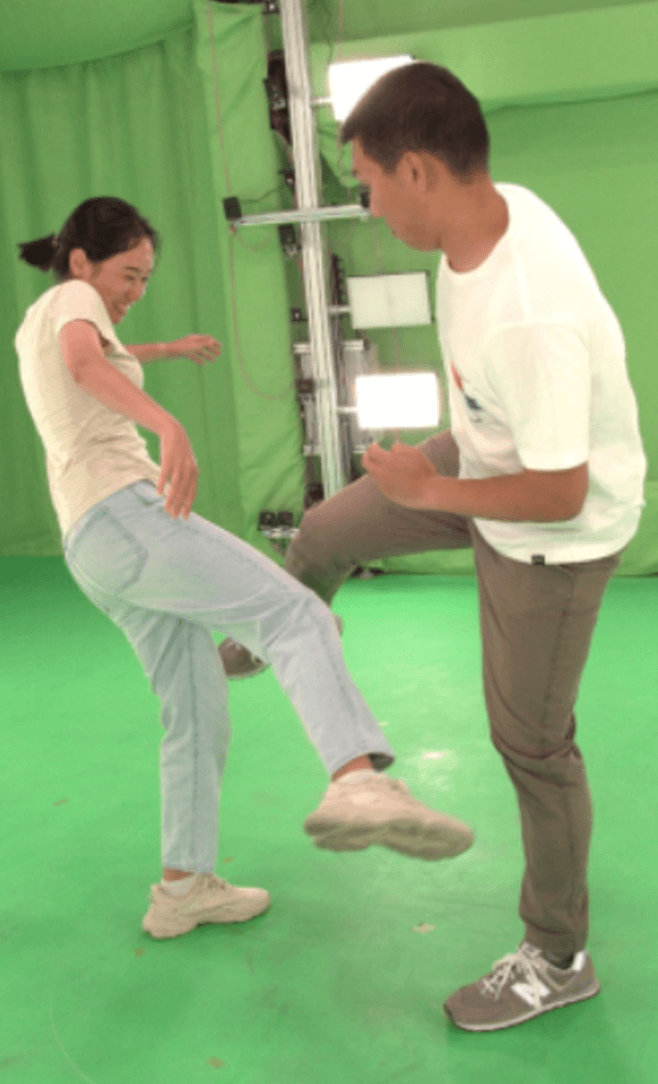}
        \captionsetup{font=scriptsize}
        \caption*{Input}
    \end{subfigure}
    \hspace{-0.2cm}
    \begin{subfigure}{60pt}
        \centering
        \includegraphics[width=60pt, height=98.8pt]{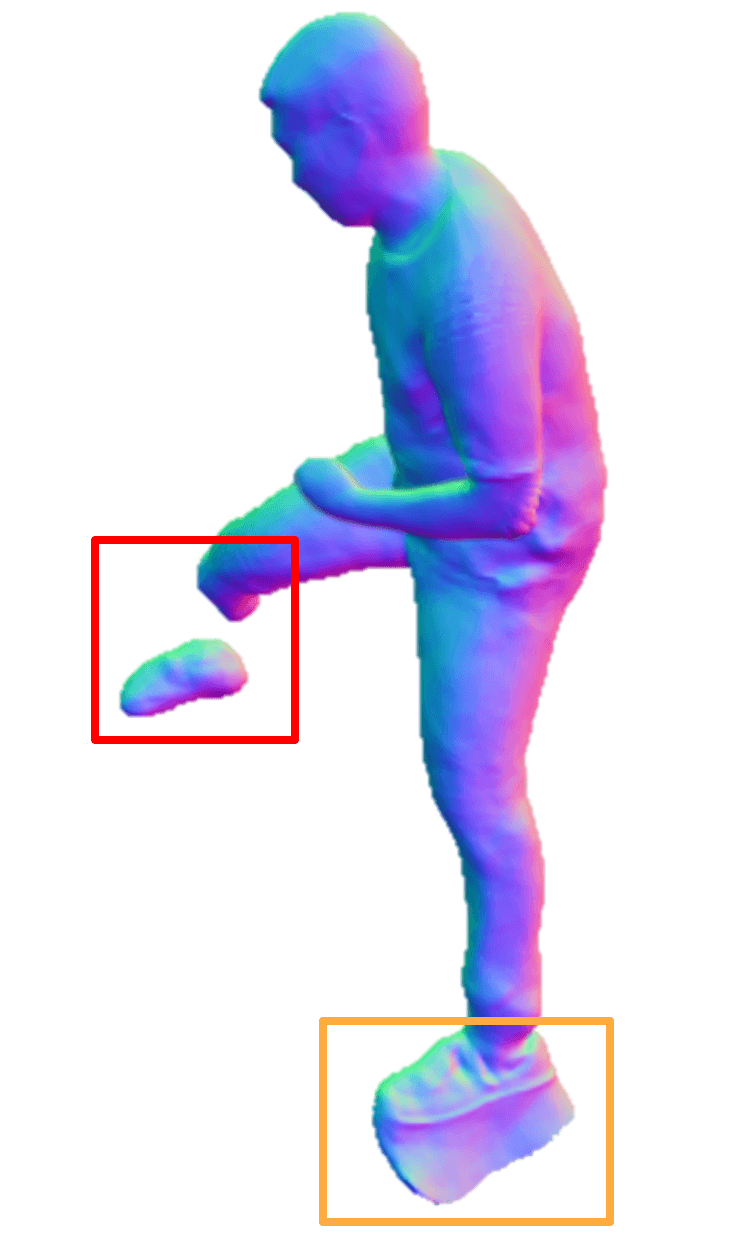}
        \captionsetup{font=scriptsize}
        \caption*{Volume rendering}
    \end{subfigure}
    \hspace{-0.2cm}
    \begin{subfigure}{60pt}
        \centering
        \includegraphics[width=60pt, height=98.8pt]{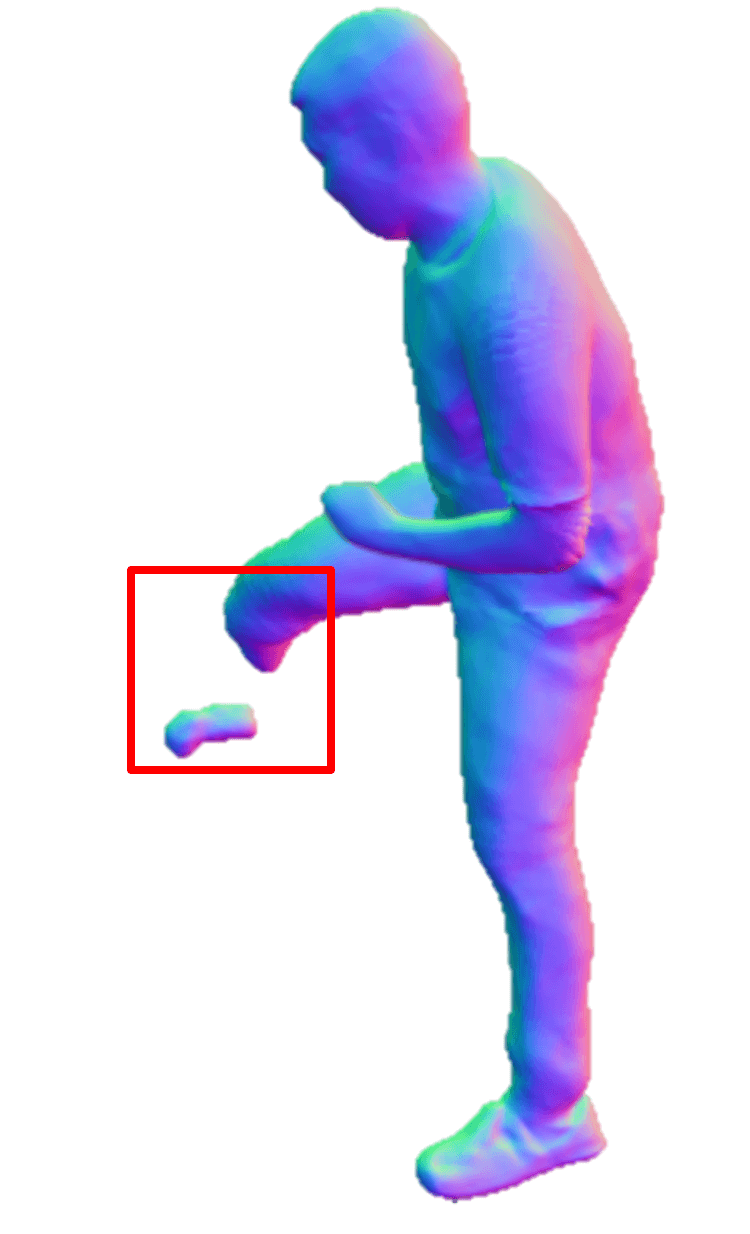}
        \captionsetup{font=scriptsize}
        \caption*{+ Prog. SAM}
    \end{subfigure}
    \hspace{-0.2cm}
    \begin{subfigure}{60pt}
        \centering
        \includegraphics[width=60pt, height=98.8pt]{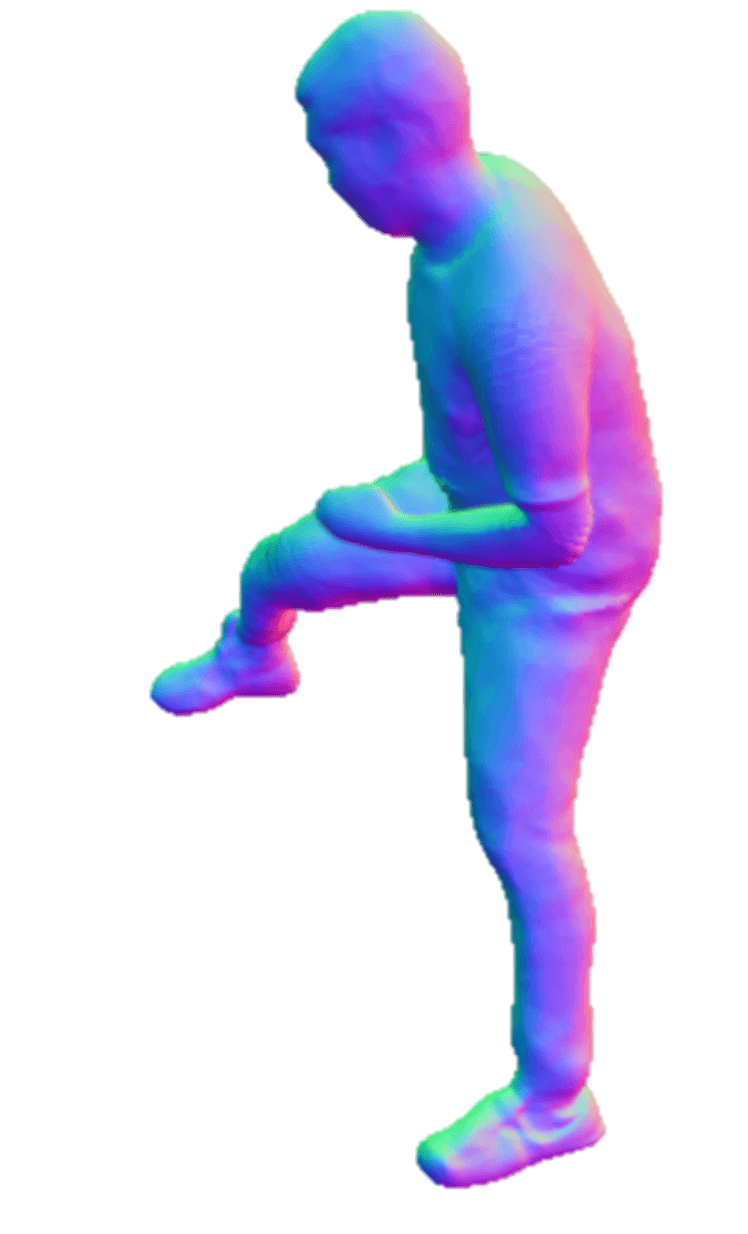}
        \captionsetup{font=scriptsize}
        \caption*{+ Conf.-guided OPT}
    \end{subfigure}
    \caption{\textbf{Qualitative ablation studies.} Our progressive prompting strategy provides robust instance segmentation supervision and eliminates the noises caused by the environmental dynamic effects. The confidence-guided optimization further improves the reconstruction results and maintains complete human bodies.}
    \vspace{-0.4cm}
    \label{fig:ablation_vis}
\end{figure}

}

\newcommand{\figureProgressiveSAM}{
\begin{figure}[t]
    \begin{subfigure}{60pt}
        \centering

        \adjincludegraphics[width=60pt, height=98.8pt]{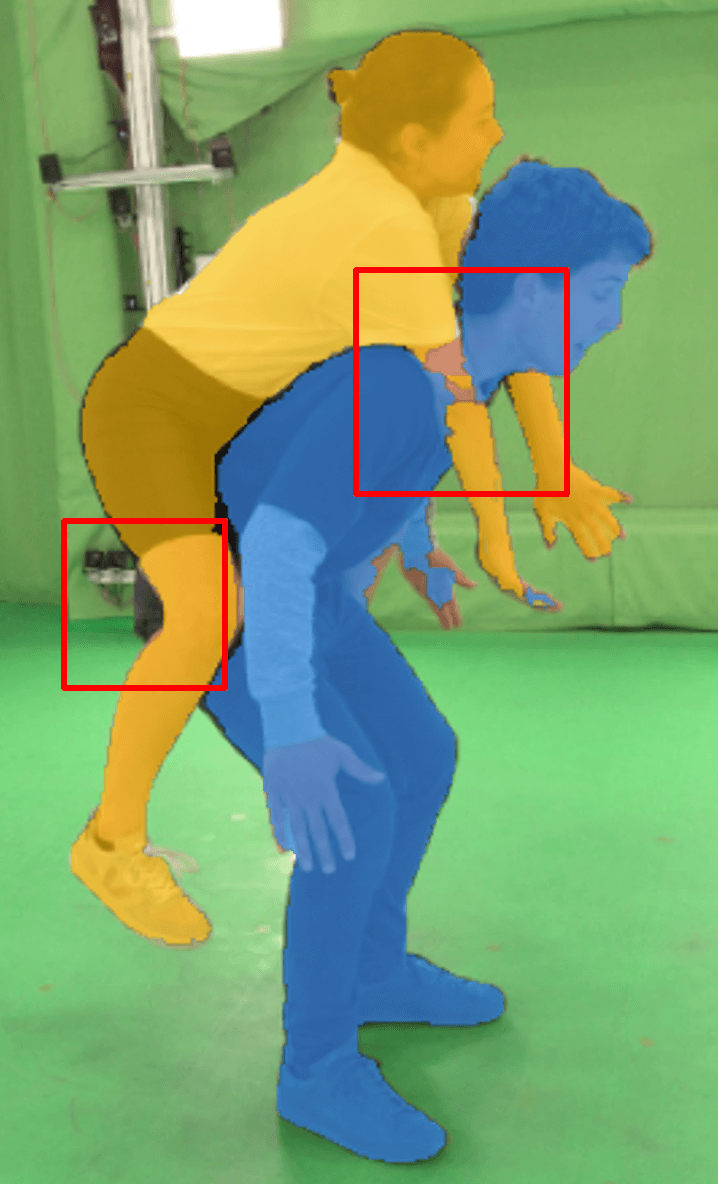}

        \caption*{Init. SAM}
    \end{subfigure}
    \hspace{-0.2cm}
    \begin{subfigure}{60pt}
        \centering
        \adjincludegraphics[width=60pt, height=98.8pt]{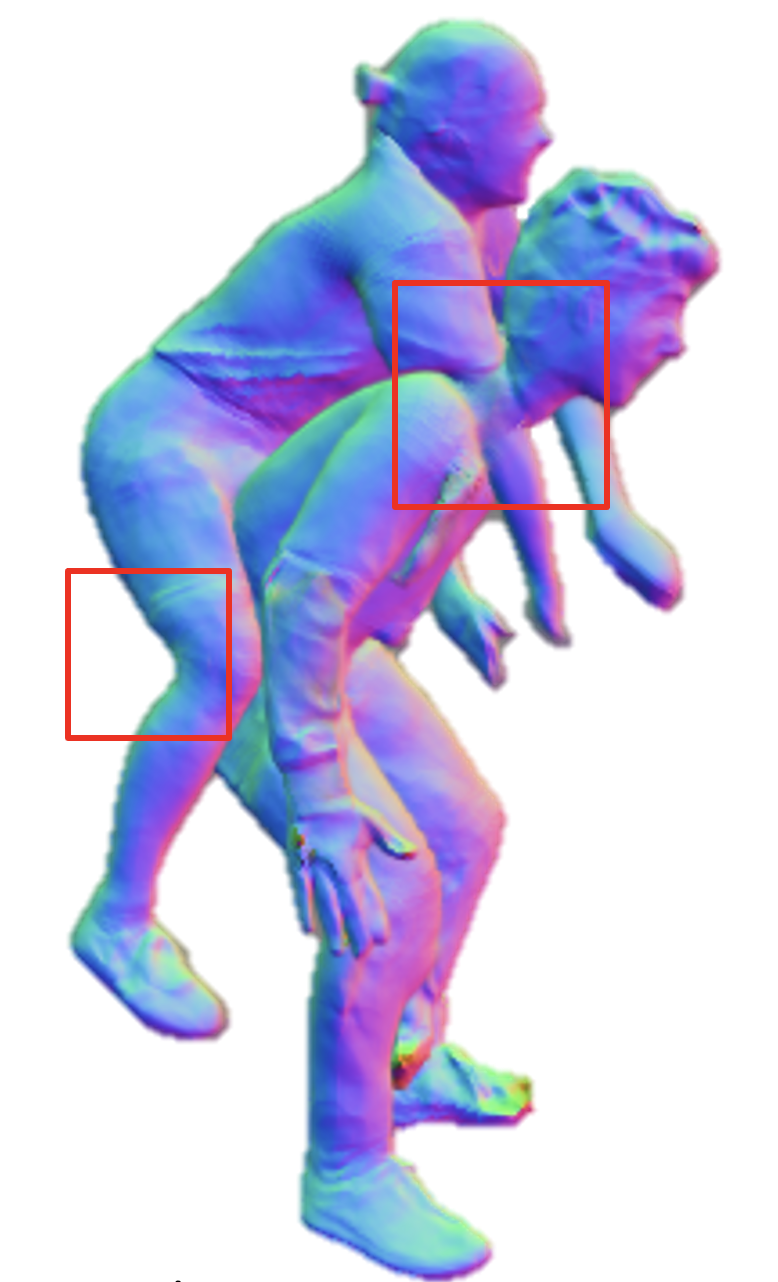}

        \caption*{Init. Recon}
    \end{subfigure}
    \hspace{-0.2cm}
    \begin{subfigure}{60pt}
        \centering
        \adjincludegraphics[width=60pt, height=98.8pt]{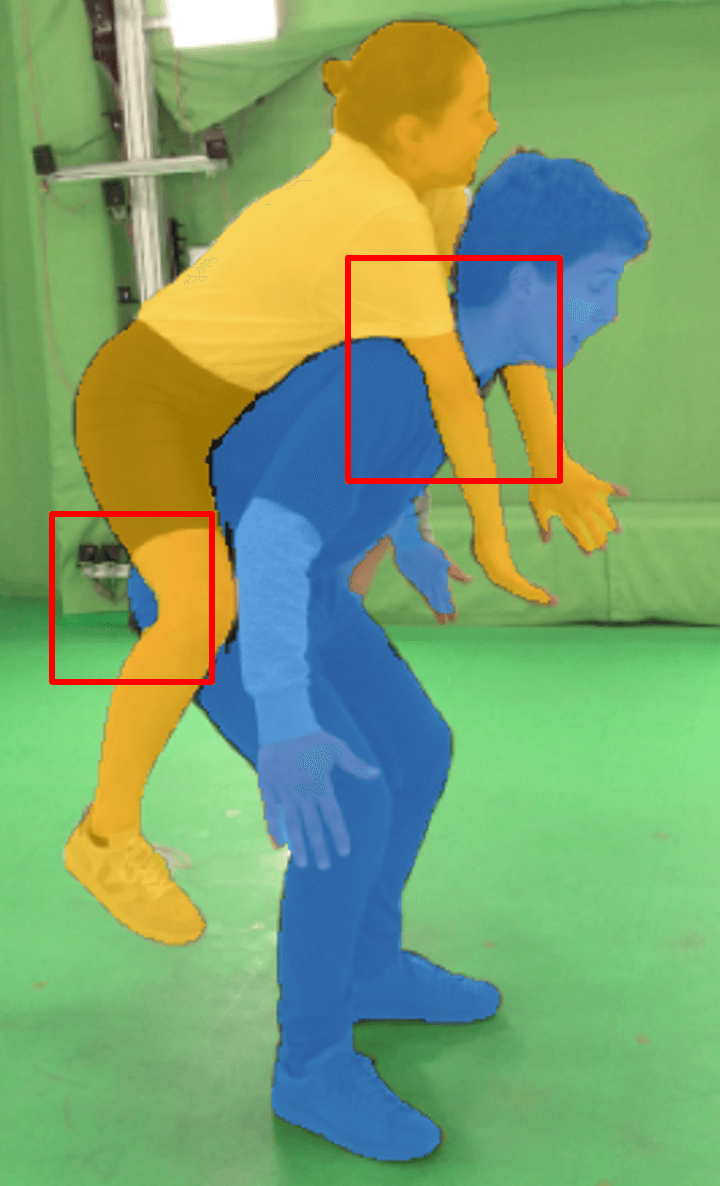}

        \caption*{Prog. SAM}
    \end{subfigure}
    \hspace{-0.2cm}
    \begin{subfigure}{60pt}
        \centering
        \adjincludegraphics[width=60pt, height=98.8pt]{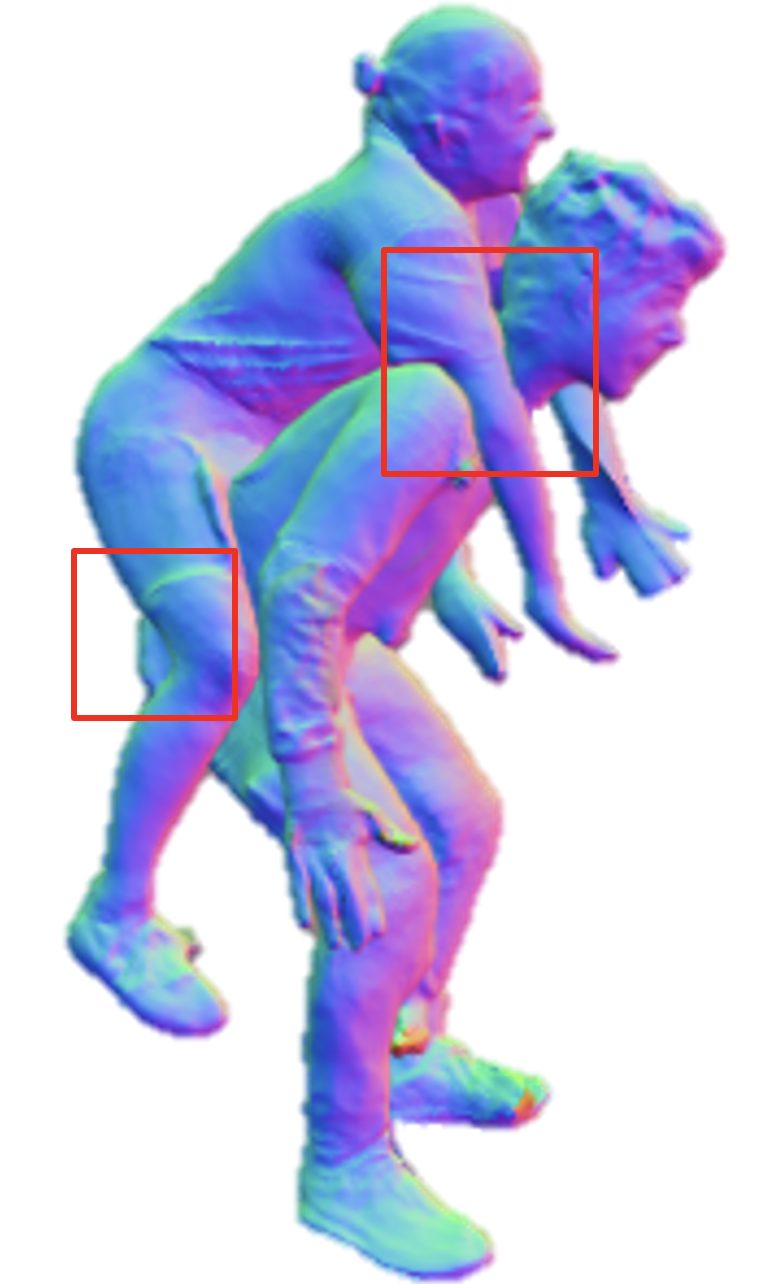}

        \caption*{Prog. Recon.}
    \end{subfigure}
    \caption{\textbf{Qualitative instance segmentation comparison.} Progressive prompting strategy provides more robust and fine-grained instance segmentation supervision compared to the initial SAM outputs, leading to higher quality of reconstructions.}
    \vspace{-0.4cm}
    \label{fig:seg}
\end{figure}
}

\newcommand{\figureNVS}{
\begin{figure}[t]
    \begin{subfigure}{60pt}
        \centering
        \adjincludegraphics[width=60pt, height=98.8pt, trim={0 {.025\height} 0 {.025\height}},clip]{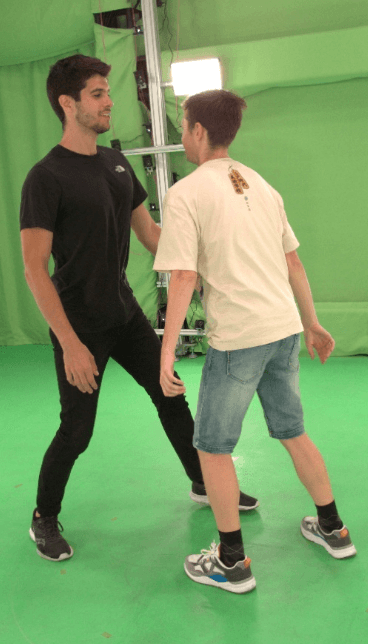}
        \caption*{Input}
    \end{subfigure}
    \hspace{-0.2cm}
    \begin{subfigure}{60pt}
        \centering
        \adjincludegraphics[width=60pt, height=98.8pt, trim={0 {.025\height} 0 {.025\height}},clip]{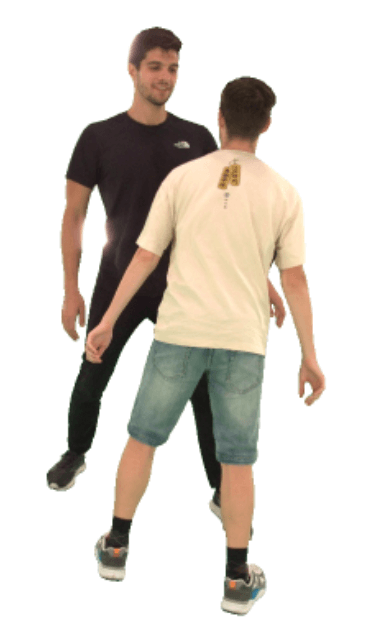}
        \caption*{Reference}
    \end{subfigure}
    \hspace{-0.2cm}
    \begin{subfigure}{60pt}
        \centering
        \adjincludegraphics[width=60pt, height=98.8pt, trim={0 {.025\height} 0 {.025\height}},clip]{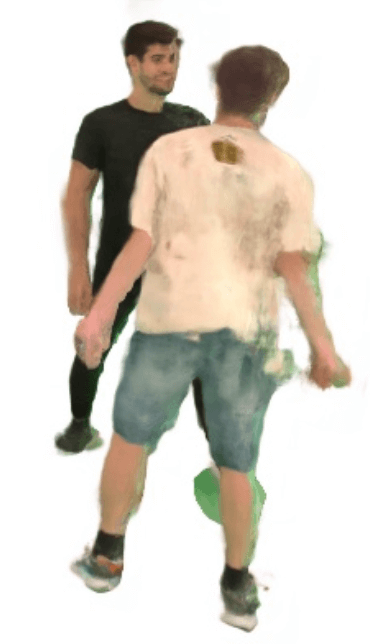}
        \caption*{Shuai et al.}
    \end{subfigure}
    \hspace{-0.2cm}
    \begin{subfigure}{60pt}
        \centering
        \adjincludegraphics[width=60pt, height=98.8pt, trim={0 {.025\height} 0 {.025\height}},clip]{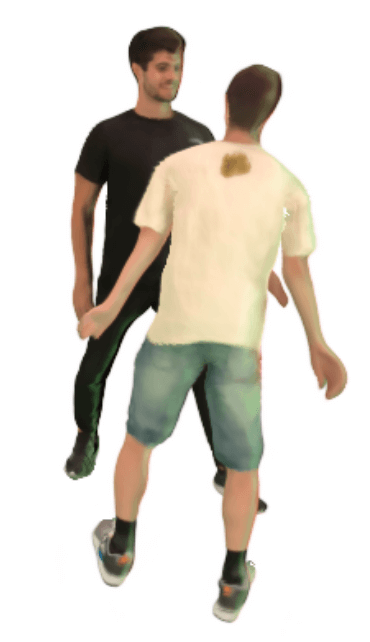}
        \caption*{Ours}
    \end{subfigure}
    \caption{\textbf{Qualitative rendering comparison.} Our method achieves more plausible renderings with sharp boundaries.}
    \vspace{-0.15cm}
    \label{fig:nvs}
\end{figure}
}

\newcommand{\figureSupNVS}{
\begin{figure}[t]
    \begin{subfigure}{60pt}
        \centering
        \adjincludegraphics[width=60pt, height=80pt, trim={{.125\width} {.132\height} {.125\width} {.132\height}},clip]{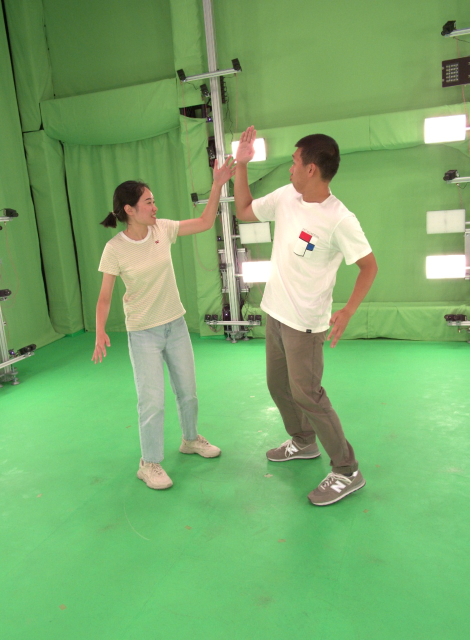}

    \end{subfigure}
    \hspace{-0.2cm}
    \begin{subfigure}{60pt}
        \centering
        \adjincludegraphics[width=60pt, height=80pt, trim={{.125\width} {.132\height} {.125\width} {.132\height}},clip]{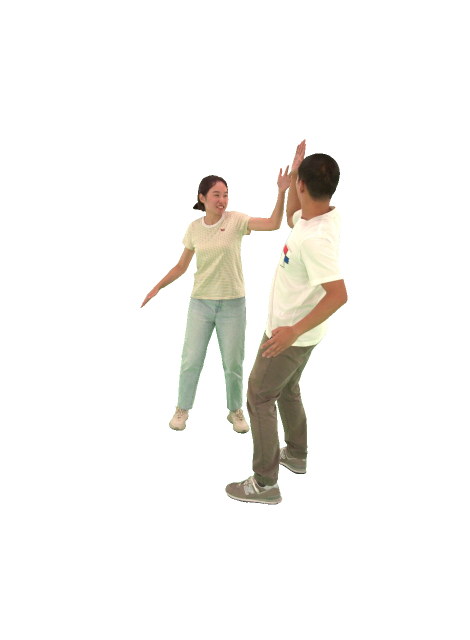}

    \end{subfigure}
    \hspace{-0.2cm}
    \begin{subfigure}{60pt}
        \centering
        \adjincludegraphics[width=60pt, height=80pt, trim={{.125\width} {.132\height} {.125\width} {.132\height}},clip]{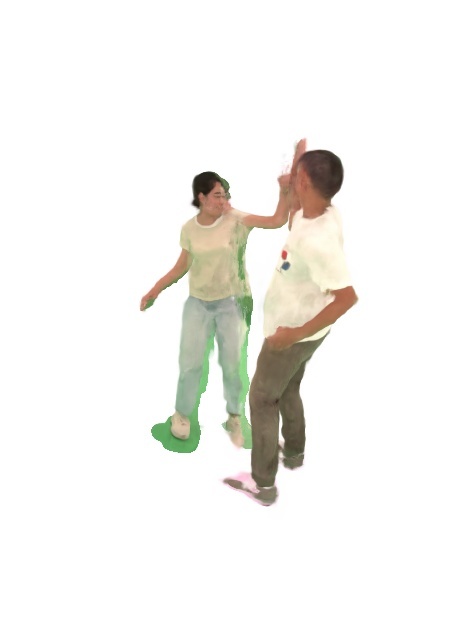}

    \end{subfigure}
    \hspace{-0.2cm}
    \begin{subfigure}{60pt}
        \centering
        \adjincludegraphics[width=60pt, height=80pt, trim={{.125\width} {.132\height} {.125\width} {.132\height}},clip]{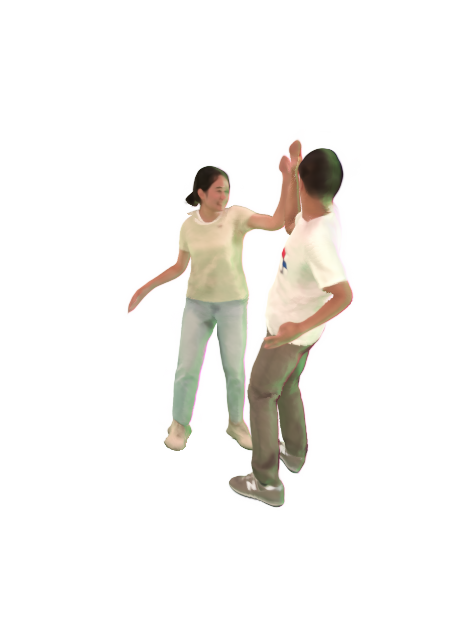}

    \end{subfigure}
    \\
    \begin{subfigure}{60pt}
        \centering
        \adjincludegraphics[width=60pt, height=80pt, trim={{.125\width} {.132\height} {.125\width} {.132\height}},clip]{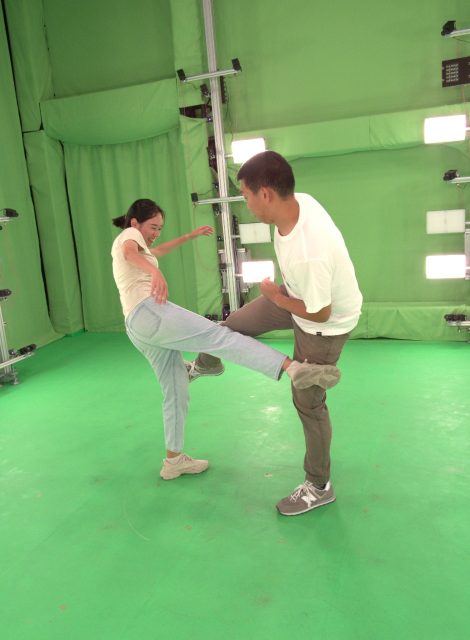}

        \caption*{Input}
    \end{subfigure}
    \hspace{-0.2cm}
    \begin{subfigure}{60pt}
        \centering
        \adjincludegraphics[width=60pt, height=80pt, trim={{.125\width} {.132\height} {.125\width} {.132\height}},clip]{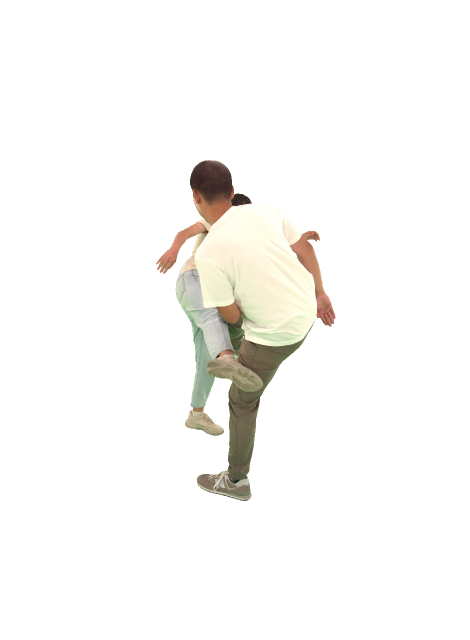}

        \caption*{Reference}
    \end{subfigure}
    \hspace{-0.2cm}
    \begin{subfigure}{60pt}
        \centering
        \adjincludegraphics[width=60pt, height=80pt, trim={{.125\width} {.132\height} {.125\width} {.132\height}},clip]{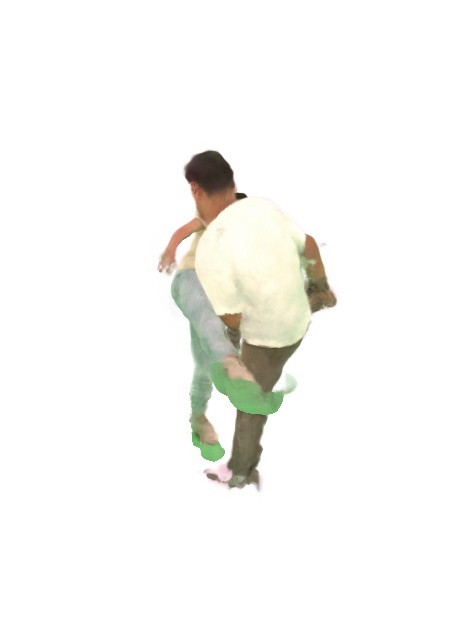}

        \caption*{Shuai et al.}
    \end{subfigure}
    \hspace{-0.2cm}
    \begin{subfigure}{60pt}
        \centering
        \adjincludegraphics[width=60pt, height=80pt, trim={{.125\width} {.132\height} {.125\width} {.132\height}},clip]{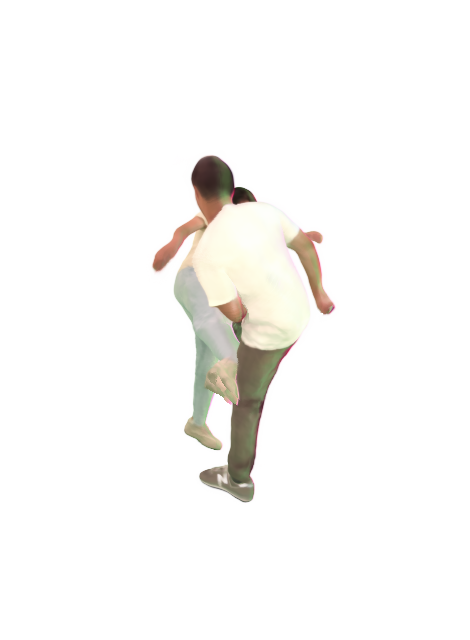}

        \caption*{Ours}
    \end{subfigure}
    \caption{\textbf{Additional qualitative rendering comparison.} Our method achieves more plausible renderings with sharp boundaries.}
    \label{fig:sup_nvs}
\end{figure}
}

\newcommand{\figurePose}{

\begin{figure*}[ht]
    
    \begin{subfigure}{70pt}
        \centering
        
        \includegraphics[width=70pt, height=95.5pt]{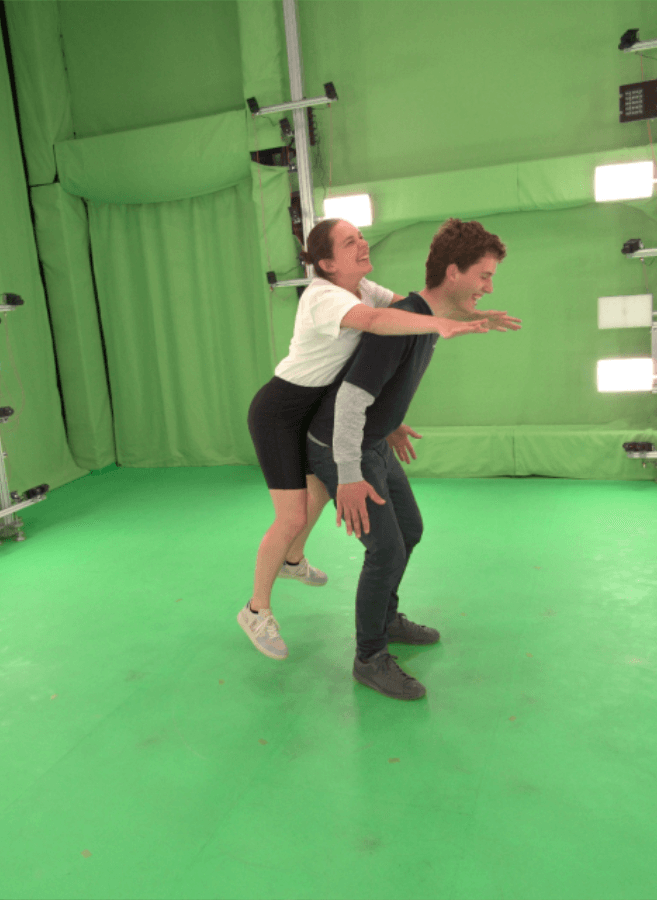}
        
    \end{subfigure}
    \hspace{-0.2cm}
    \begin{subfigure}{70pt}
        \centering
        \includegraphics[width=70pt, height=95.5pt]{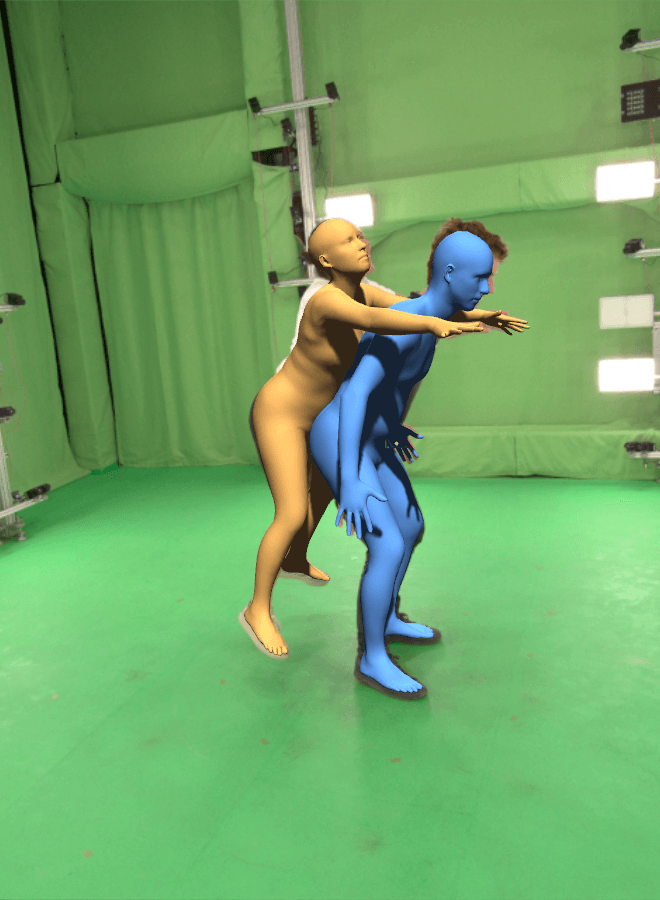}
        
    \end{subfigure}
    \hspace{-0.2cm}
    \begin{subfigure}{70pt}
        \centering
        \includegraphics[width=70pt, height=95.5pt]{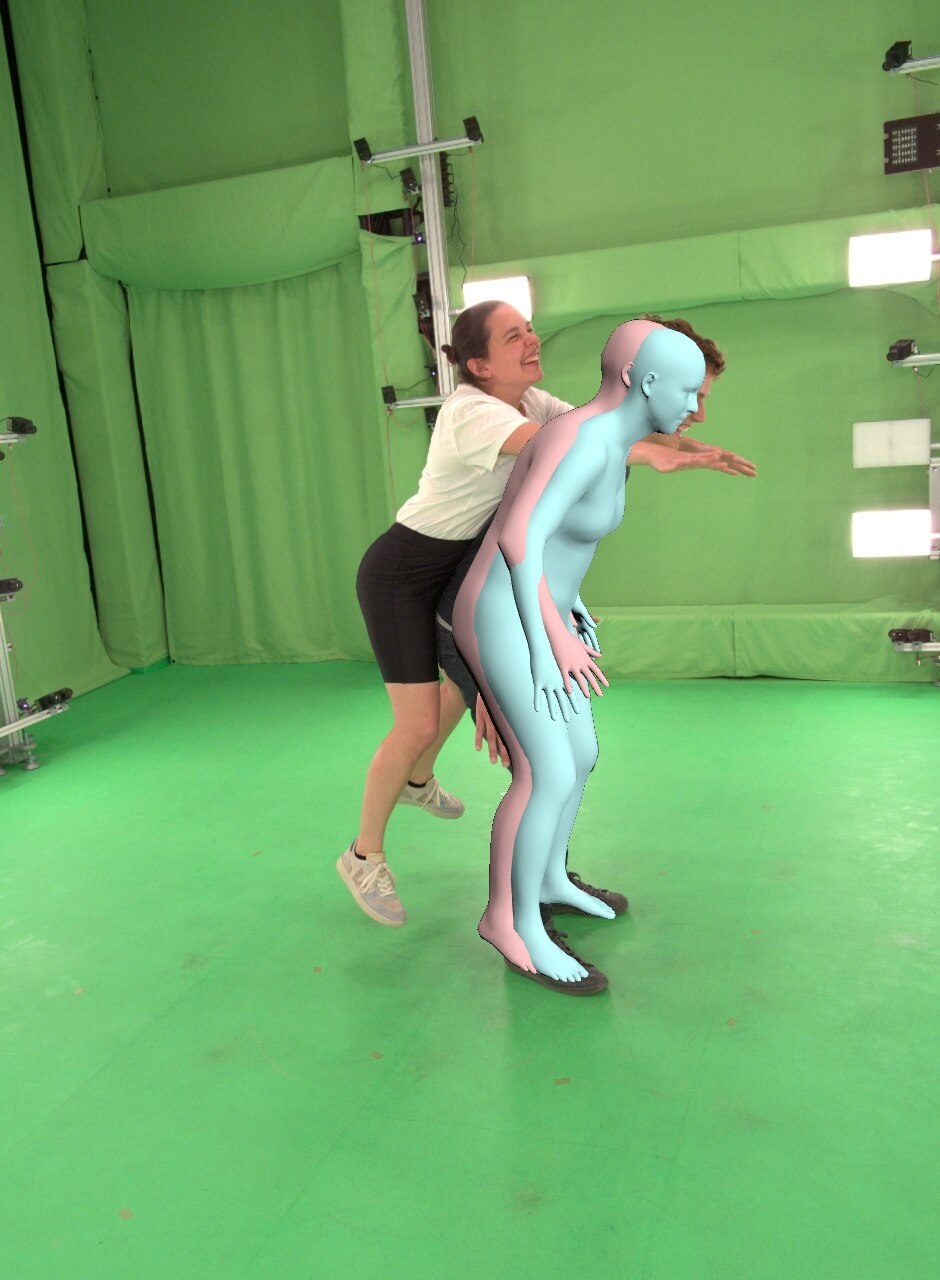}
        
    \end{subfigure}
    \hspace{-0.2cm}
    \begin{subfigure}{70pt}
        \centering
        \includegraphics[width=70pt, height=95.5pt]{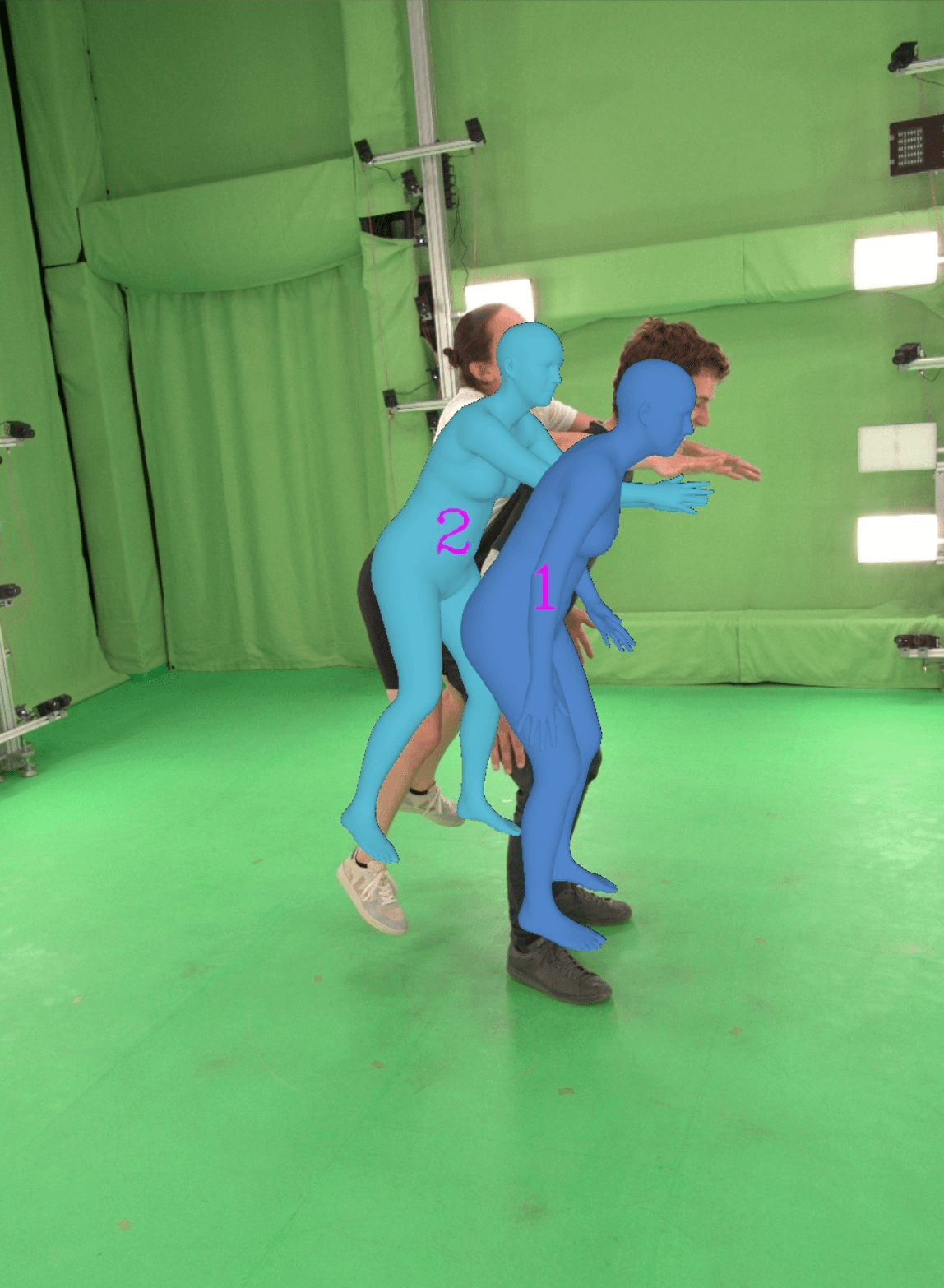}
        
    \end{subfigure}
    \hspace{-0.2cm}
    \begin{subfigure}{70pt}
        \centering
        \includegraphics[width=70pt, height=95.5pt]{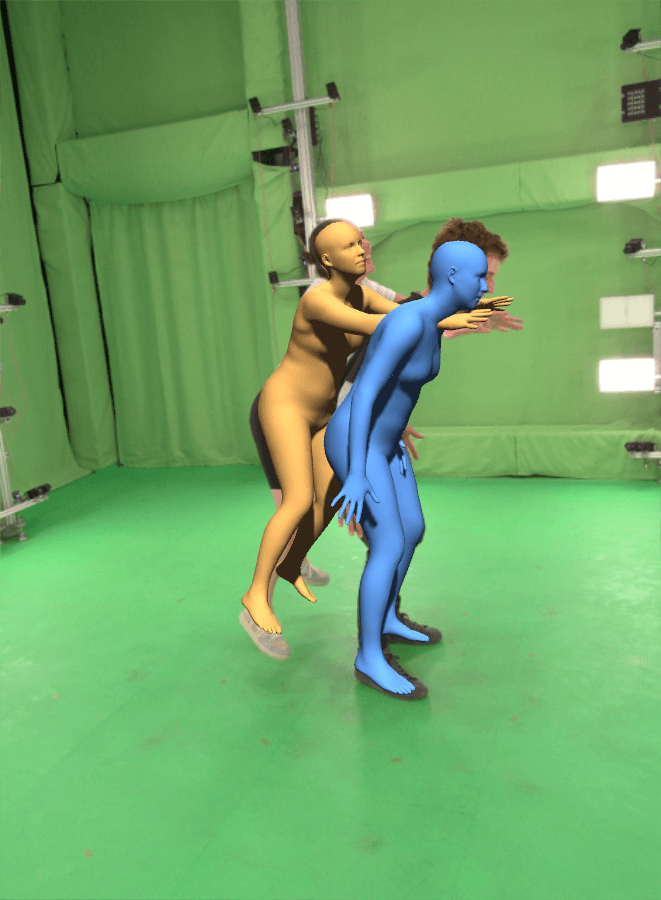}
        
    \end{subfigure}
    \hspace{-0.2cm}
    \begin{subfigure}{70pt}
        \centering
        \includegraphics[width=70pt, height=95.5pt]{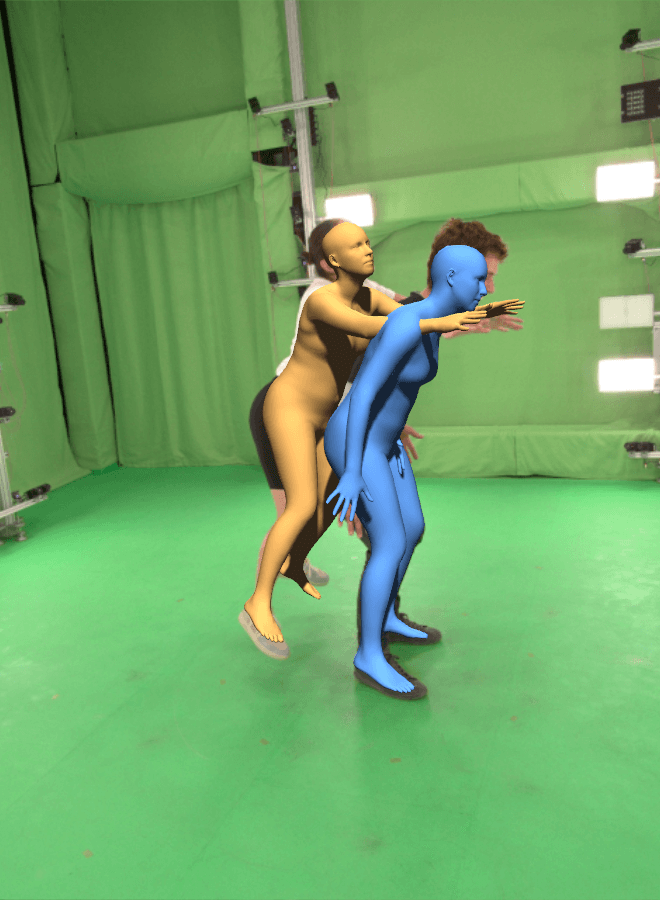}
        
    \end{subfigure}
    \hspace{-0.2cm}
    \begin{subfigure}{70pt}
        \centering
        \includegraphics[width=70pt, height=95.5pt]{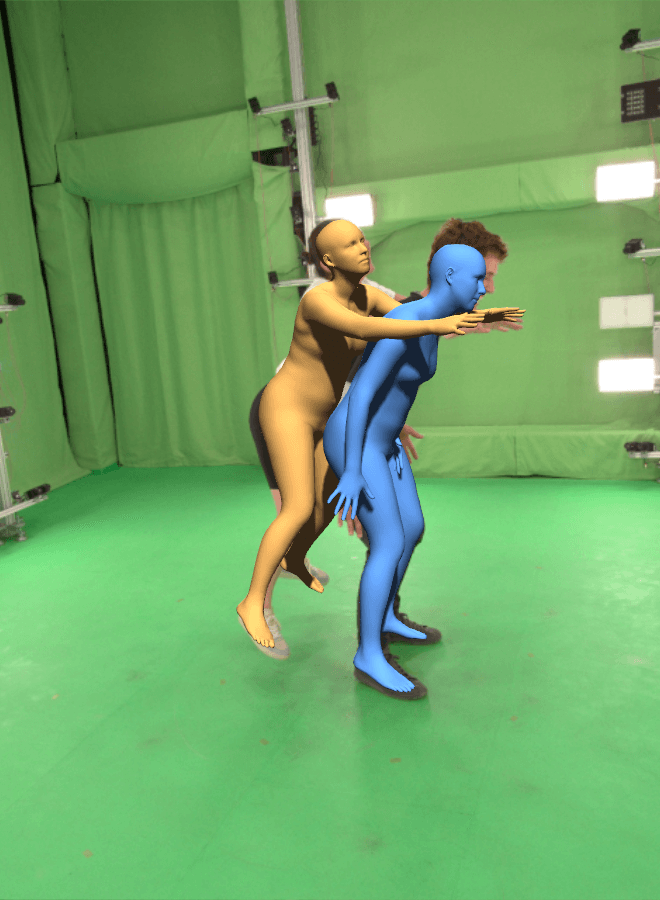}
        
    \end{subfigure}
    \vspace{-0.27cm}
    \\    
    \begin{subfigure}{70pt}
        \centering
        \includegraphics[width=70pt, height=95.5pt]{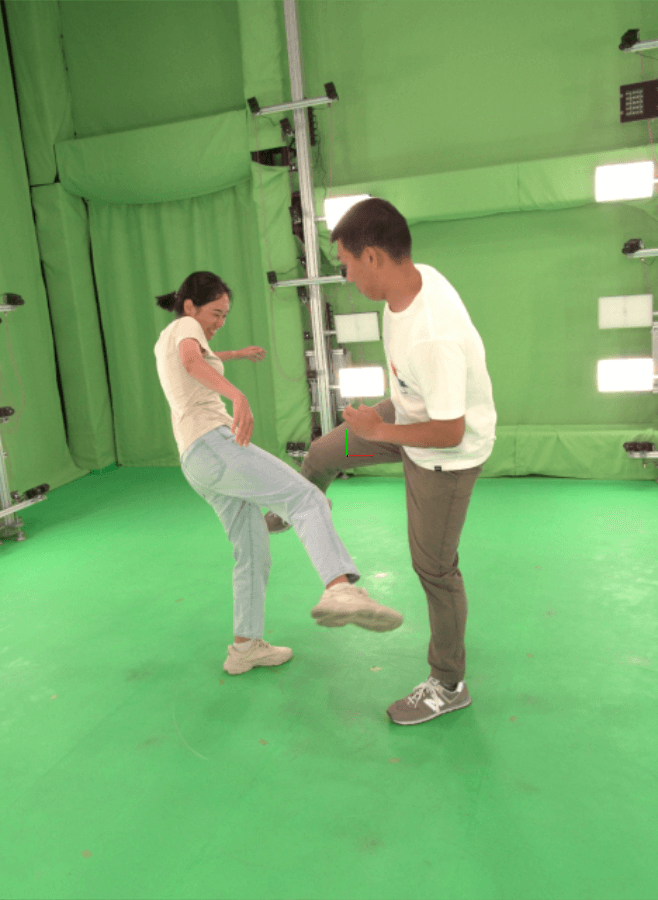}
        \captionsetup{font=scriptsize}
        \caption*{Image}
    \end{subfigure}
    \hspace{-0.2cm}
    \begin{subfigure}{70pt}
        \centering
        \includegraphics[width=70pt, height=95.5pt]{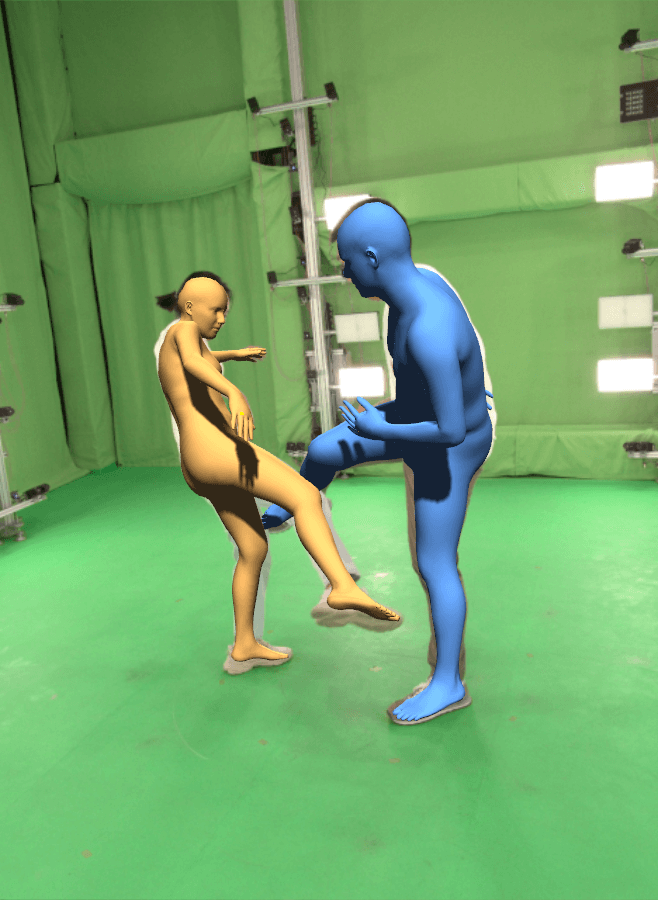}
        \captionsetup{font=scriptsize}
        \caption*{GT}
    \end{subfigure}
    \hspace{-0.2cm}
    \begin{subfigure}{70pt}
        \centering
        \includegraphics[width=70pt, height=95.5pt]{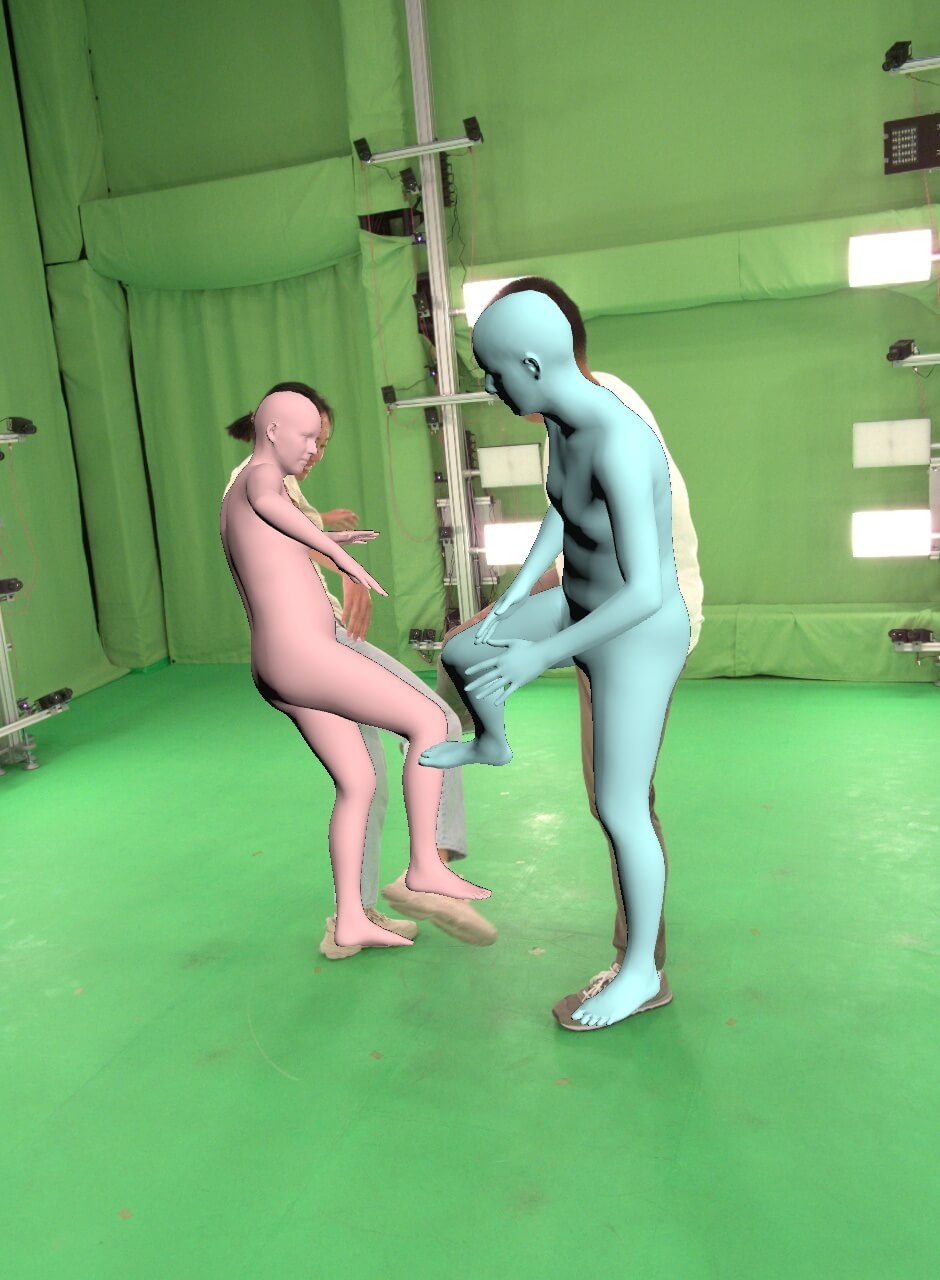}
        \captionsetup{font=scriptsize}
        \caption*{CLIFF}
    \end{subfigure}
    \hspace{-0.2cm}
    \begin{subfigure}{70pt}
        \centering
        \includegraphics[width=70pt, height=95.5pt]{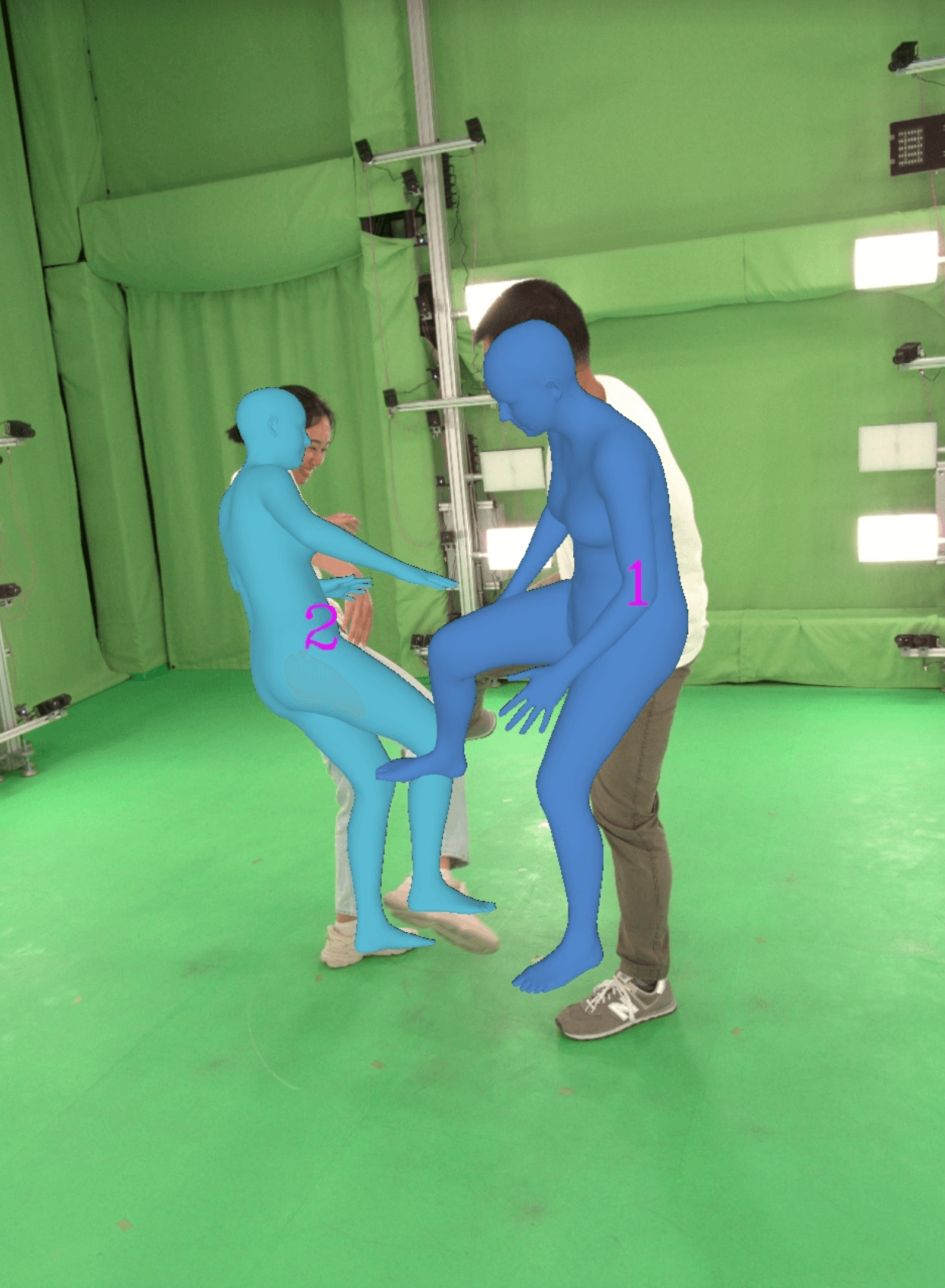}
        \captionsetup{font=scriptsize}
        \caption*{TRACE}
    \end{subfigure}
    \hspace{-0.2cm}
    \begin{subfigure}{70pt}
        \centering
        \includegraphics[width=70pt, height=95.5pt]{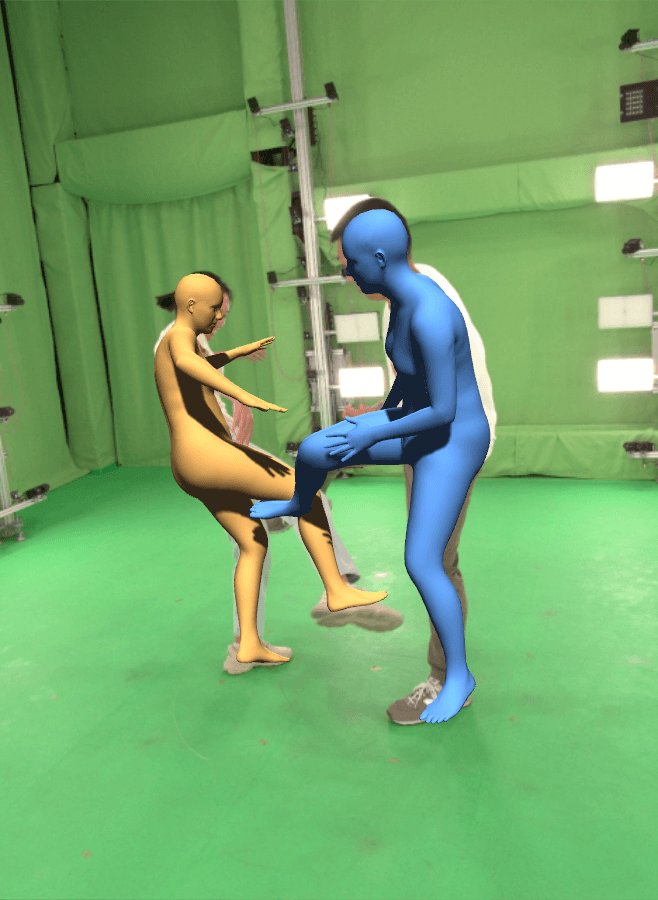}
        \captionsetup{font=scriptsize}
        \caption*{Ours Init.}
    \end{subfigure}
    \hspace{-0.2cm}
    \begin{subfigure}{70pt}
        \centering
        \includegraphics[width=70pt, height=95.5pt]{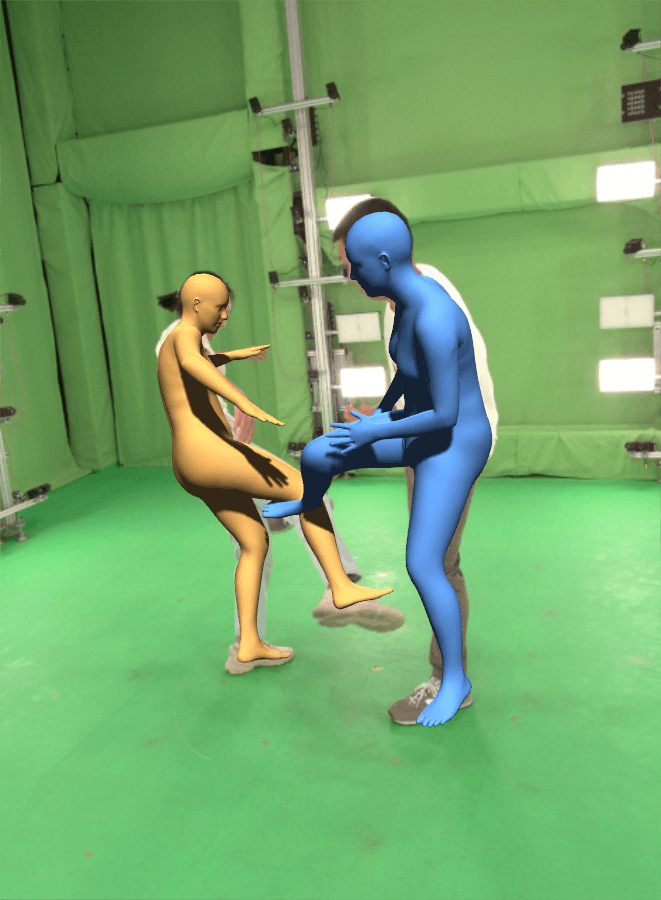}
        \captionsetup{font=scriptsize}
        \caption*{Joint OPT}
    \end{subfigure}
    \hspace{-0.2cm}
    \begin{subfigure}{70pt}
        \centering
        \includegraphics[width=70pt, height=95.5pt]{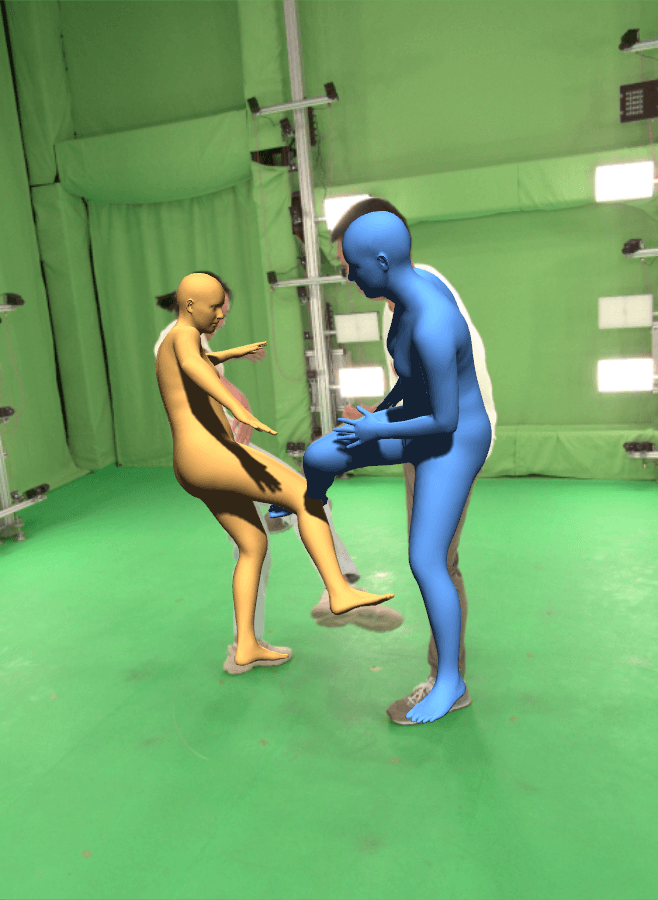}
        \captionsetup{font=scriptsize}
        \caption*{Ours}
    \end{subfigure}
    \caption{\textbf{Qualitative pose estimation comparisons.} CLIFF, TRACE, and our pose initialization all fail to estimate the correct spatial arrangement of the close interacted pairs on the Hi4D dataset. Simply optimizing both pose and shape jointly during training doesn't help to refine inaccurate pose estimates. In contrast, our confidence-guided alternating optimization performs effectively in correcting implausible human poses and spatial arrangement (e.g. the interpenetration between the arm and the body, and the wrong depth order of legs).}
    \vspace{-0.4cm}
    \label{fig:pose_comp}
\end{figure*}

}

\newcommand{\figureSupMMM}{
\begin{figure}[t]
    \begin{subfigure}{80pt}
        \centering
        
        \adjincludegraphics[width=80pt, height=107pt]{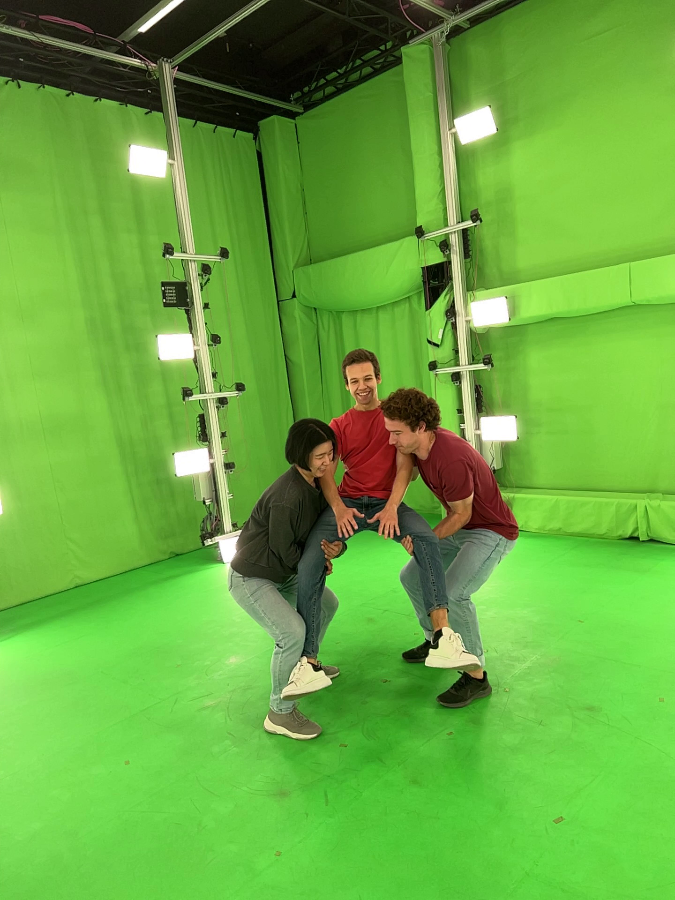}

    \end{subfigure}
    \hspace{-0.2cm}
    \begin{subfigure}{80pt}
        \centering
        \adjincludegraphics[width=80pt, height=107pt]{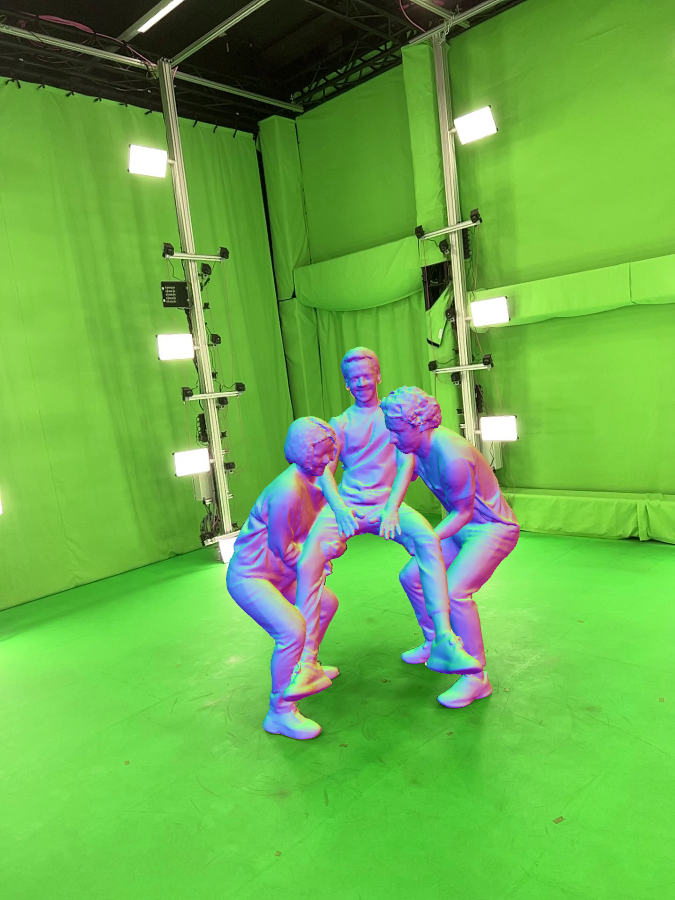}

    \end{subfigure}
    \hspace{-0.2cm}
    \begin{subfigure}{80pt}
        \centering
        \adjincludegraphics[width=80pt, height=107pt]{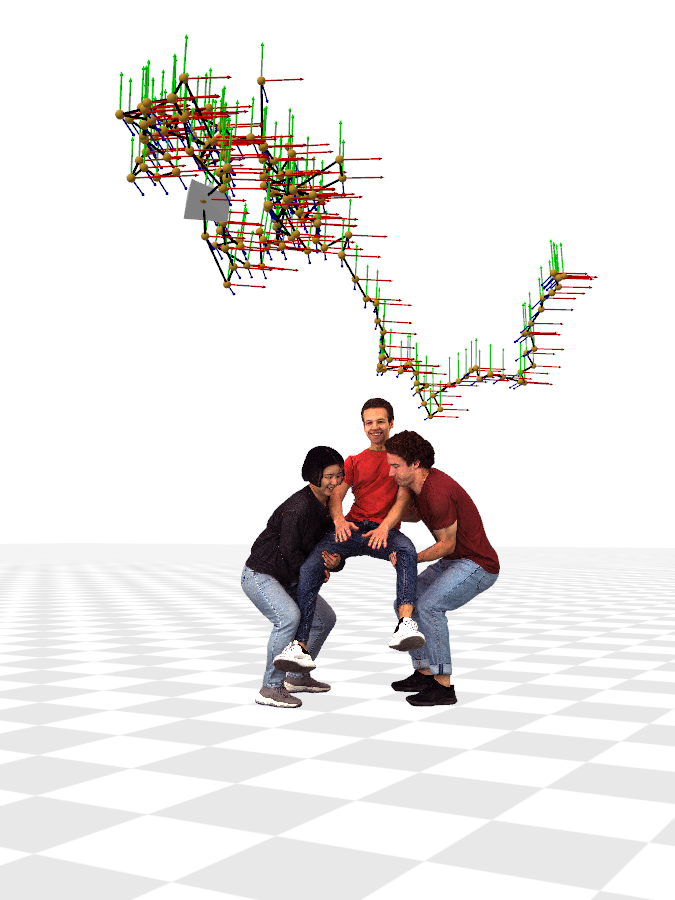}

    \end{subfigure}
    \\
    \begin{subfigure}{80pt}
        \centering
        
        \adjincludegraphics[width=80pt, height=107pt]{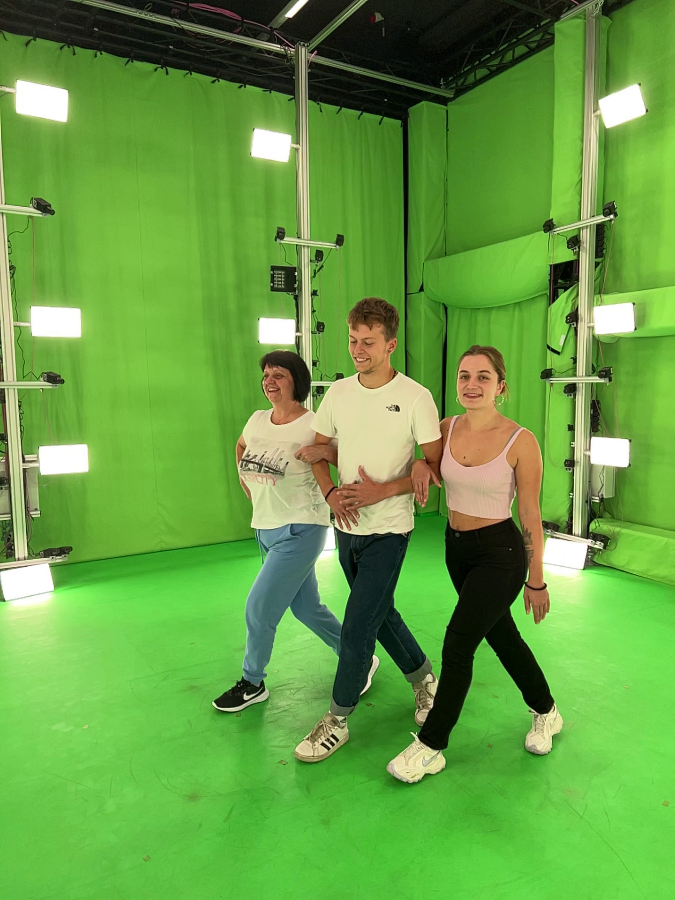}

    \end{subfigure}
    \hspace{-0.2cm}
    \begin{subfigure}{80pt}
        \centering
        \adjincludegraphics[width=80pt, height=107pt]{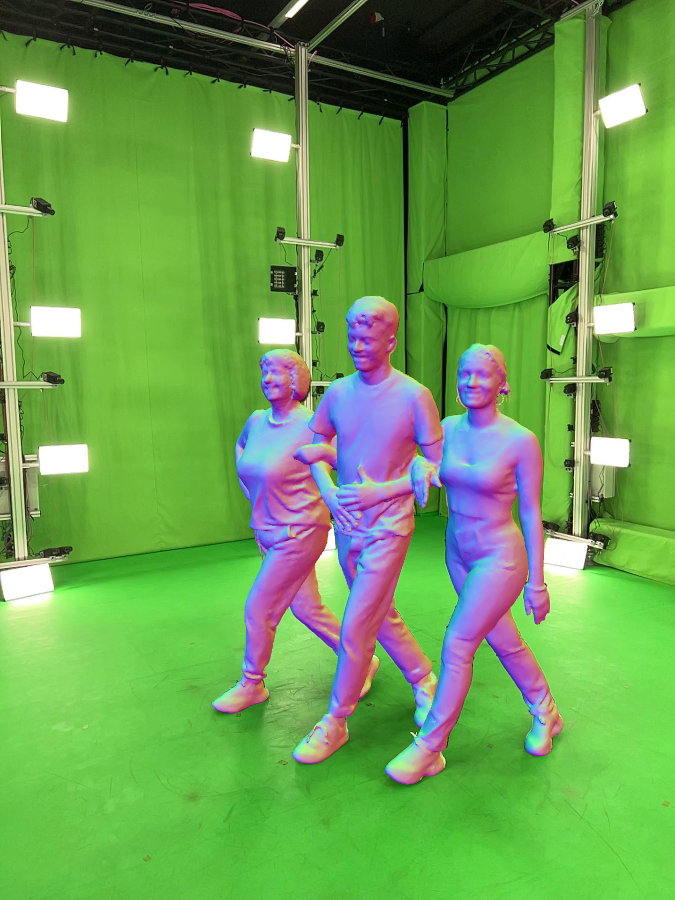}

    \end{subfigure}
    \hspace{-0.2cm}
    \begin{subfigure}{80pt}
        \centering
        \adjincludegraphics[width=80pt, height=107pt]{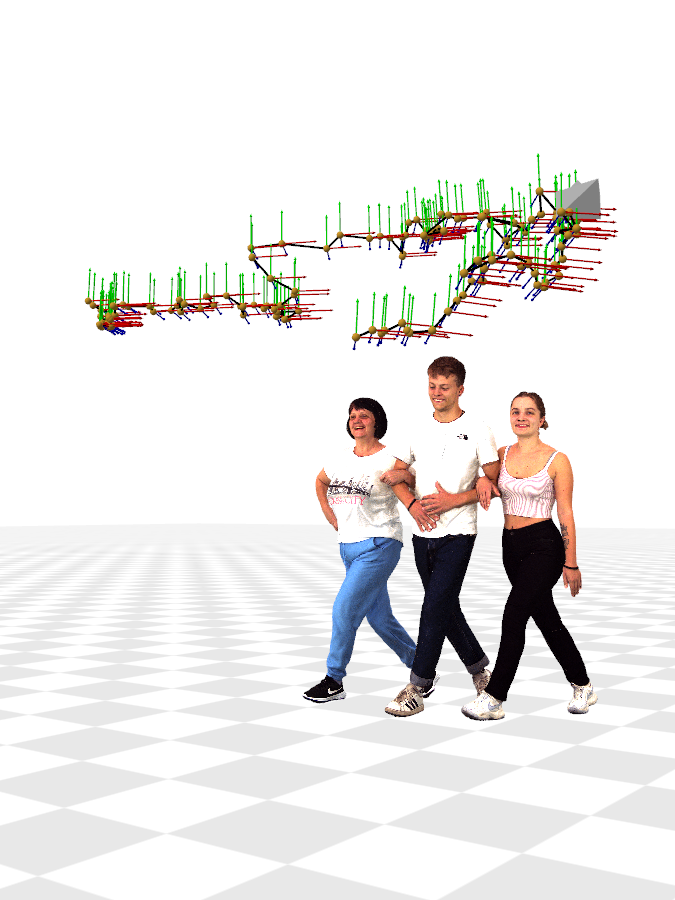}

    \end{subfigure}
    \\
    \begin{subfigure}{80pt}
        \centering
        
        \adjincludegraphics[width=80pt, height=107pt]{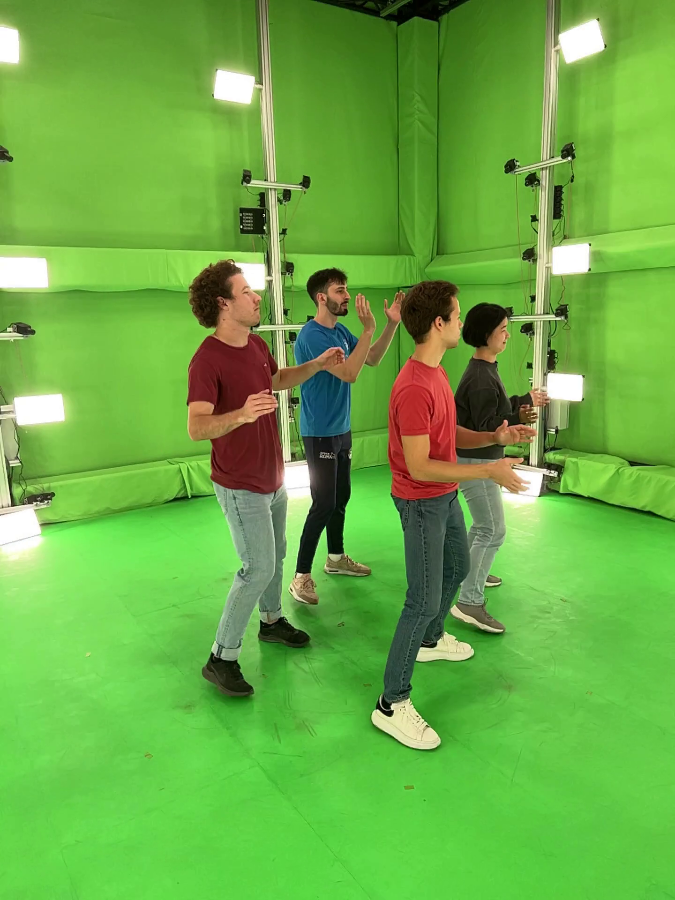}

        \caption*{Image}
    \end{subfigure}
    \hspace{-0.2cm}
    \begin{subfigure}{80pt}
        \centering
        \adjincludegraphics[width=80pt, height=107pt]{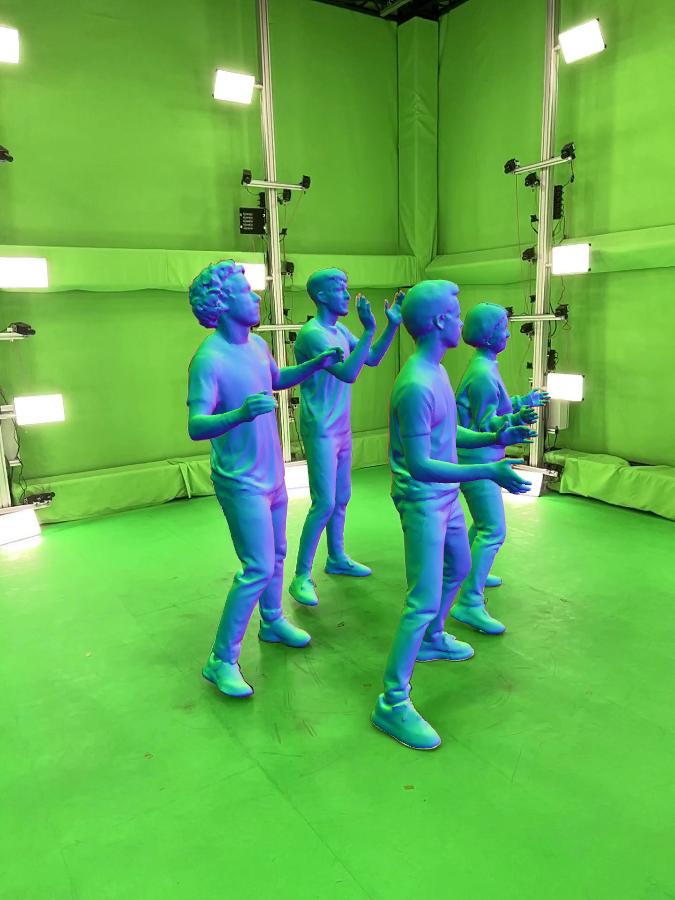}

        \caption*{GT Meshes}
    \end{subfigure}
    \hspace{-0.2cm}
    \begin{subfigure}{80pt}
        \centering
        \adjincludegraphics[width=80pt, height=107pt]{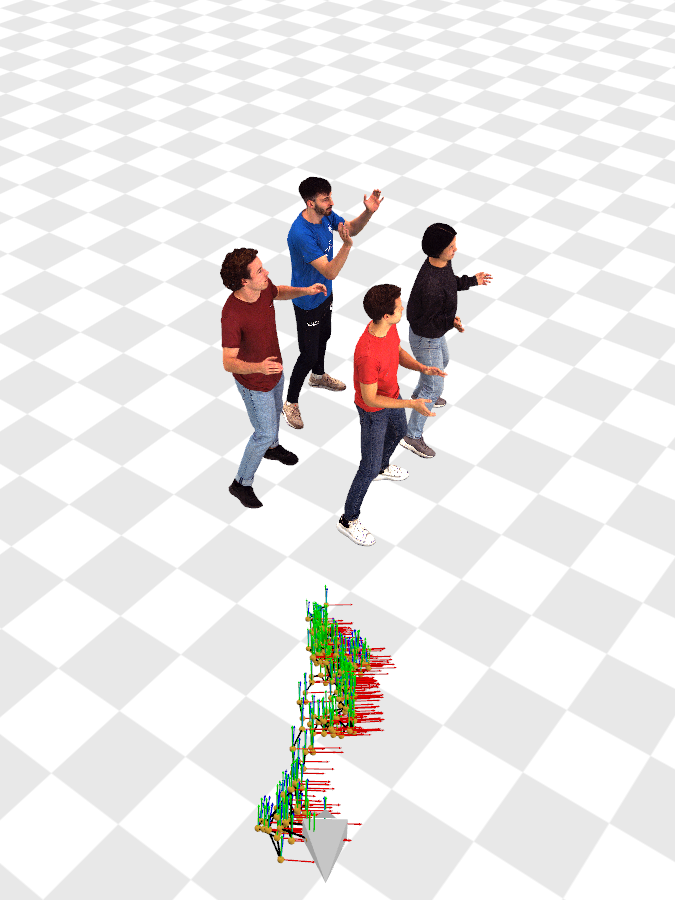}

        \caption*{Camera Trajectories}
    \end{subfigure}
    \caption{\textbf{MMM Dataset.} We show the captured image, ground-truth meshes, and ground-truth camera trajectories of our MMM dataset.}
    \label{fig:sup_mmm}
\end{figure}
}

\newcommand{\figureSupVis}{

\begin{figure*}[t]
    \centering
    \includegraphics[width=\textwidth,trim=0 0 0 0,clip]{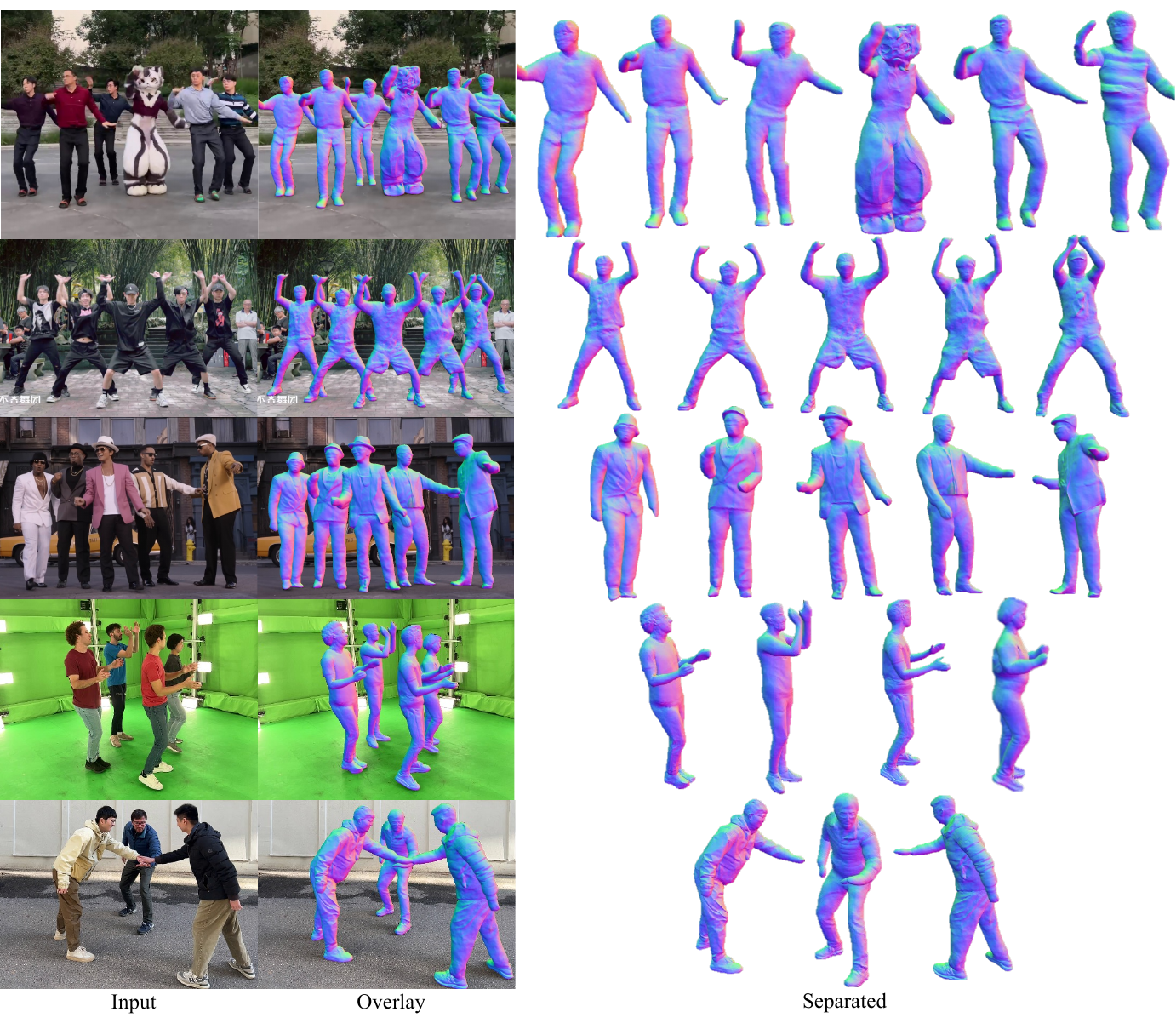}
    \caption{\textbf{Additional qualitative results.} 
    Our method MultiPly generalizes to various people with different human shapes and miscellaneous clothing styles and performs robustly against different levels of occlusions, close human interaction, and environmental visual complexities.
    }
    \label{fig:sup_vis}

\end{figure*}

}

\newcommand{\figureSupSeg}{
\begin{figure}[t]
    \begin{subfigure}{80pt}
        \centering
        
        \adjincludegraphics[width=80pt, height=109pt]{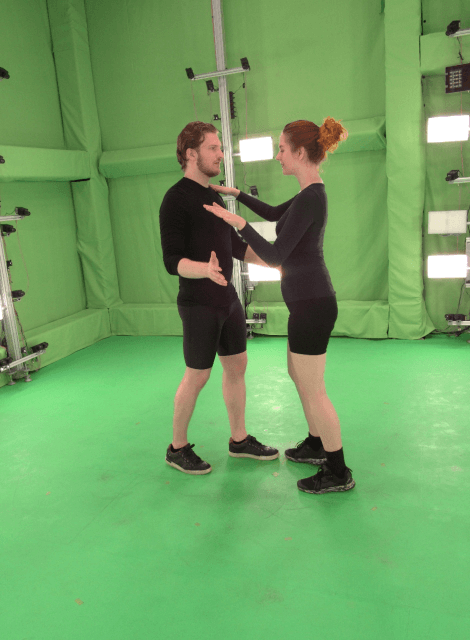}

    \end{subfigure}
    \hspace{-0.2cm}
    \begin{subfigure}{80pt}
        \centering
        \adjincludegraphics[width=80pt, height=109pt]{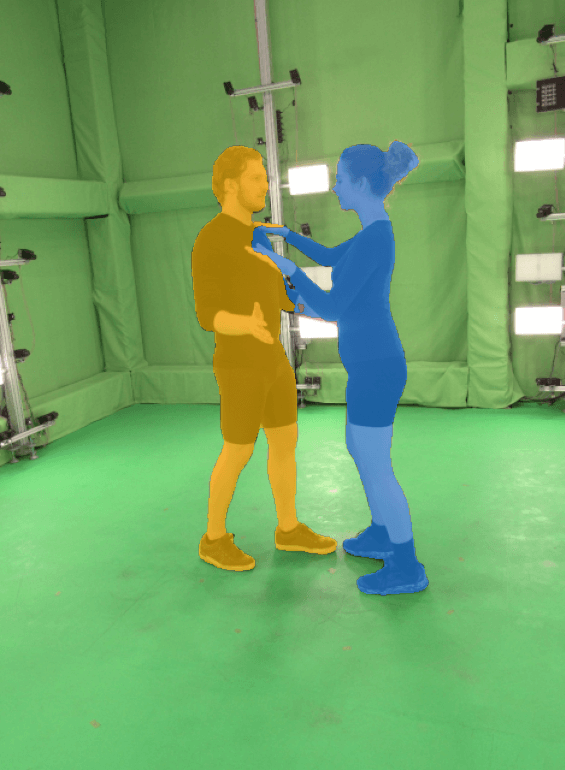}

    \end{subfigure}
    \hspace{-0.2cm}
    \begin{subfigure}{80pt}
        \centering
        \adjincludegraphics[width=80pt, height=109pt]{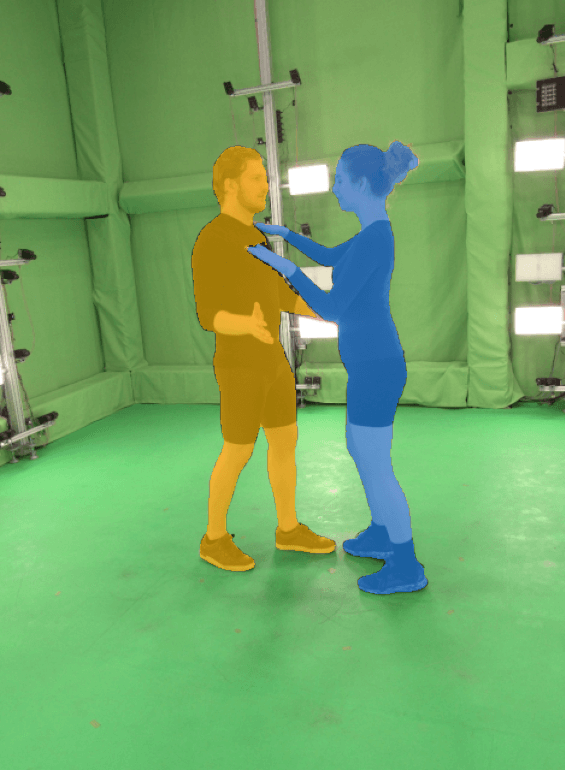}

    \end{subfigure}
    \\
    \begin{subfigure}{80pt}
        \centering
        
        \adjincludegraphics[width=80pt, height=109pt]{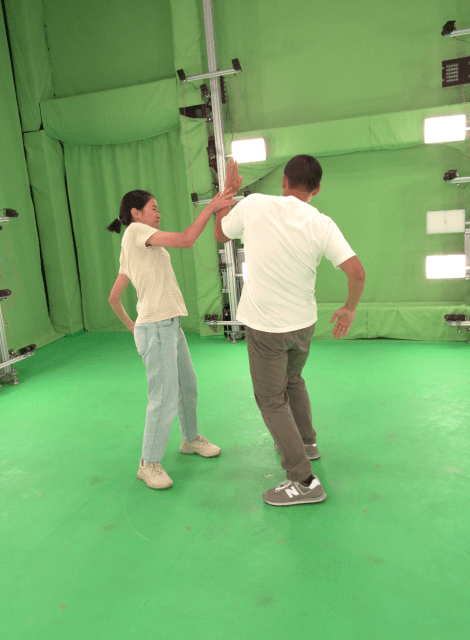}

        \caption*{Image}
    \end{subfigure}
    \hspace{-0.2cm}
    \begin{subfigure}{80pt}
        \centering
        \adjincludegraphics[width=80pt, height=109pt]{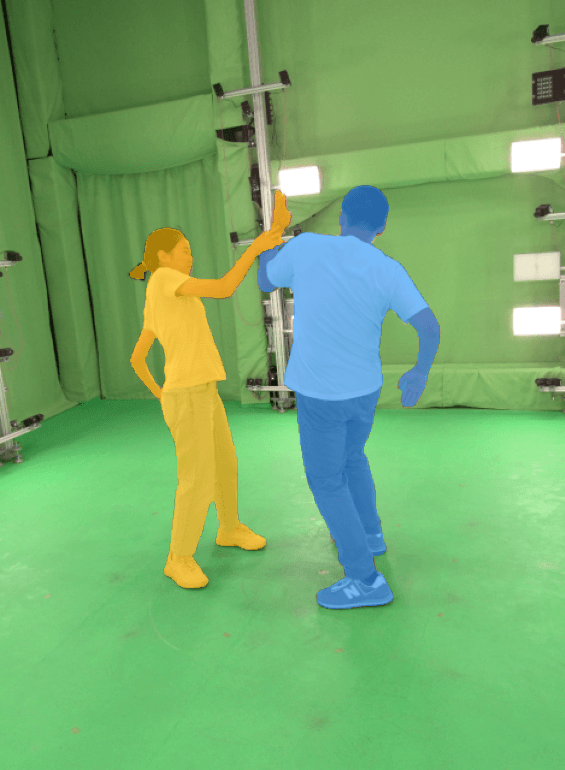}

        \caption*{SCHP}
    \end{subfigure}
    \hspace{-0.2cm}
    \begin{subfigure}{80pt}
        \centering
        \adjincludegraphics[width=80pt, height=109pt]{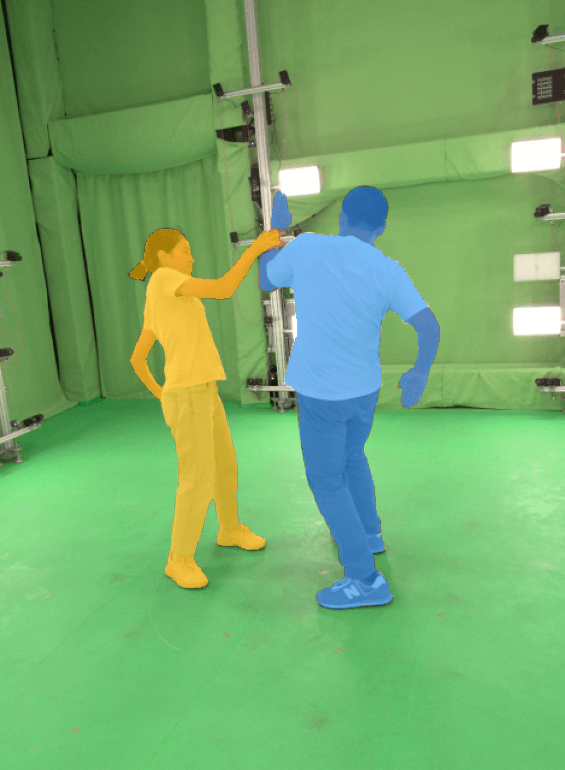}

        \caption*{Ours}
    \end{subfigure}
    \caption{\textbf{Addtional instance segmentation comparisons.} Our method produces more accurate instance segmentation masks while SCHP fails to associate the pixels to the correct subject when people closely interact.}
    \label{fig:sup_seg}
\end{figure}
}

\newcommand{\figureTri}{
\begin{figure}[t]
    \begin{subfigure}{118pt}
        \centering
        
        \adjincludegraphics[width=118pt, height=67.5pt]{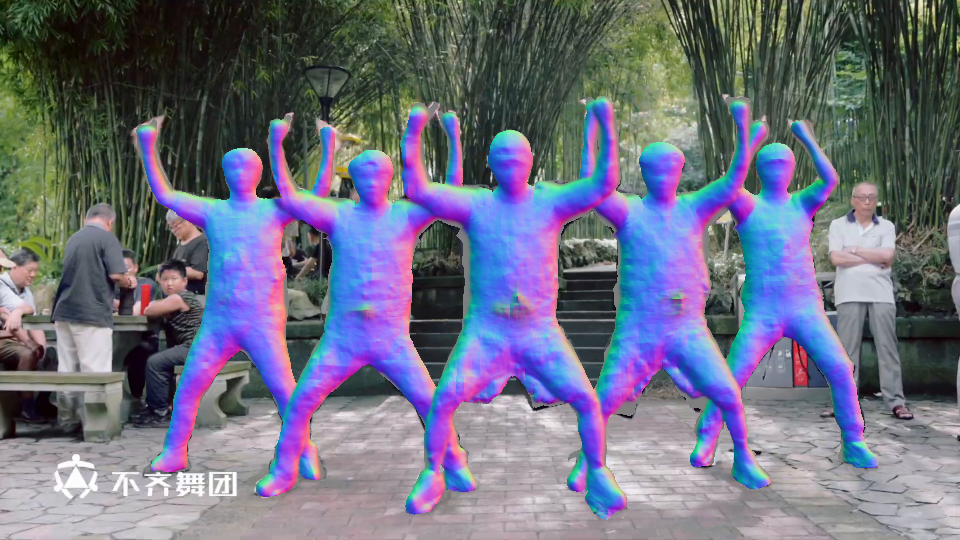}
        \caption*{Tri-plane 100 epochs}

    \end{subfigure}
    \hspace{-0.2cm}
    \begin{subfigure}{118pt}
        \centering
        \adjincludegraphics[width=120pt, height=67.5pt]{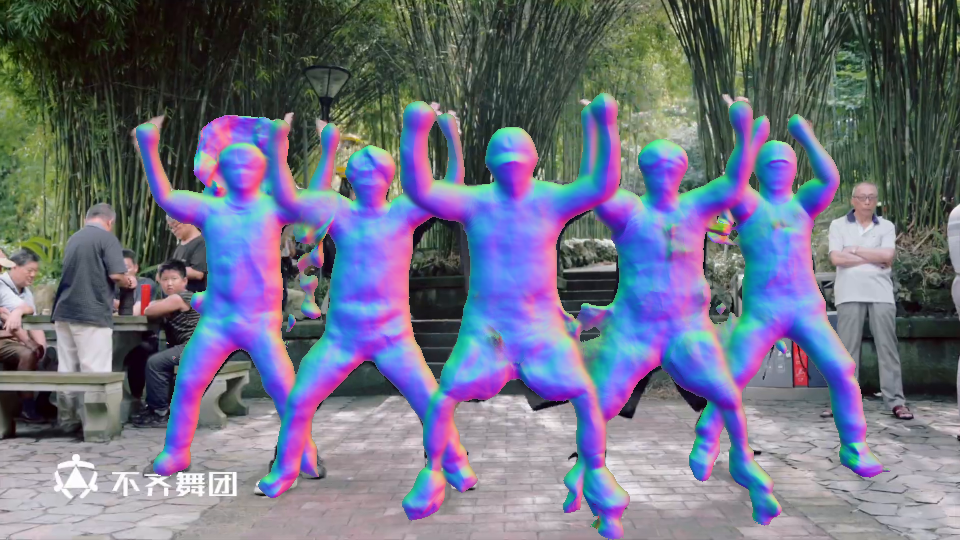}
        \caption*{MLP 100 epochs}

    \end{subfigure}
    \\
    \begin{subfigure}{118pt}
        \centering
        
        \adjincludegraphics[width=118pt, height=67.5pt]{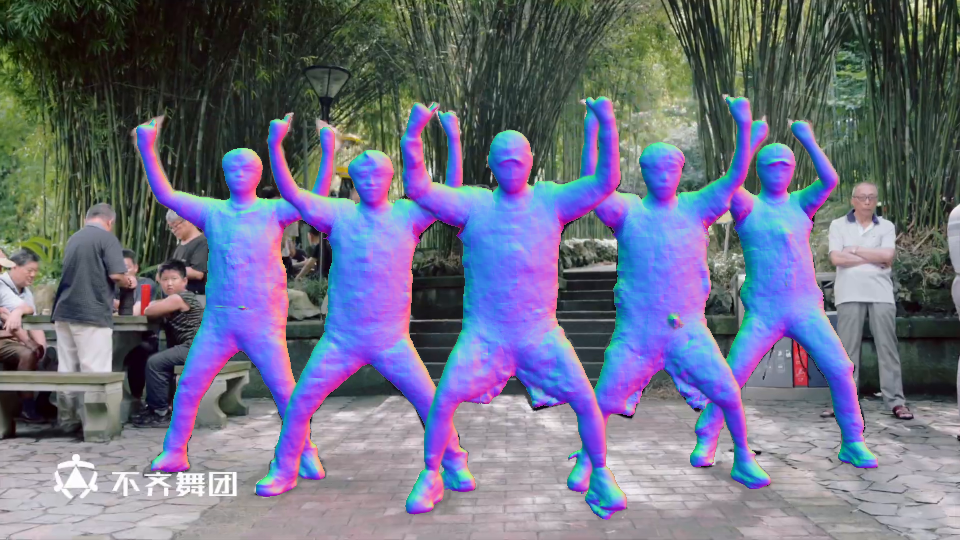}

        \caption*{Tri-plane 1000 epochs}
    \end{subfigure}
    \hspace{-0.2cm}
    \begin{subfigure}{120pt}
        \centering
        \adjincludegraphics[width=118pt, height=67.5pt]{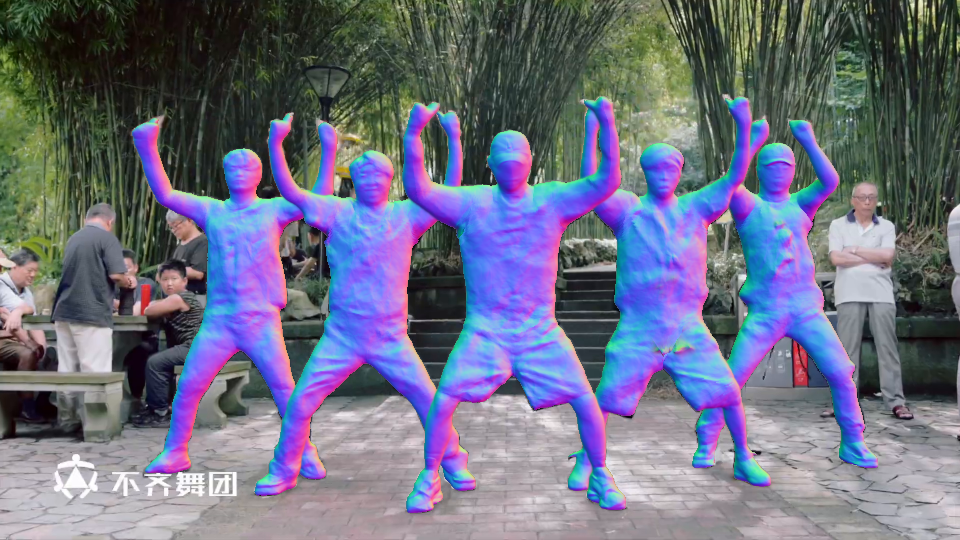}

        \caption*{MLP 1000 epochs}
    \end{subfigure}
    \caption{\textbf{Comparison between Tri-plane and MLP representation.} Tri-plane representation with a shared decoder converges faster than MLP-based human representation. However, Tri-plane representation produces less fine-grained surface details.}
    \label{fig:sup_tri}
\end{figure}
}

%% file: tables.tex
\newcommand{\tablemask}{
\begin{table}[t]
\centering
\small

\begin{tabular}{l c c c c}
\toprule
\textbf{Method} & \textbf{MPJPE}~$\downarrow$ & \textbf{MVE}~$\downarrow$ & \textbf{CD}~$\downarrow$ & \textbf{PCDR}~$\uparrow$ \\ 
\midrule
CLIFF \cite{li2022cliff}  & 85.7 & 102.1 & 351.7 & 0.606  \\ 
TRACE \cite{TRACE}  & 95.6 & 115.7 & 249.4 & 0.603  \\ 
Ours & \textbf{69.4} & \textbf{83.6} & \textbf{218.4} & \textbf{0.709}  \\
\bottomrule
\end{tabular}

\vspace{0.15cm}
\caption{\textbf{Quantitative pose estimation evaluation on Hi4D.} Our method outperforms state-of-the-art multi-person pose estimation methods on all evaluation metrics.}
\label{tab:pose}
\vspace{-0.4cm}
\end{table}
}

\newcommand{\tableAblation}{
\begin{table*}[ht]
\centering
\small

\begin{tabular}{l|cccc|cccc}
\toprule
\multirow{2}*{Metrics} & \multicolumn{4}{c|}{\textbf{Pose Estimation}} & \multicolumn{4}{c}{\textbf{Human Reconstruction}} \\ 
\cline{2-9}
~ & \textbf{MPJPE}~$\downarrow$ & \textbf{MVE}~$\downarrow$ & \textbf{CD}~$\downarrow$ & \textbf{PCDR}~$\uparrow$ & \textbf{V-IoU}~$\uparrow$ & $\mathbf{C}-\ell_{2}$~$\downarrow$ & \textbf{P2S}~$\downarrow$ &\textbf{NC}~$\uparrow$ \\ 
\midrule
Initial pose & 75.3 & 90.8 & 235.6 & 0.566 & - & - & - & - \\
\midrule
 Layer-wise volume rendering  & 71.5 & 86.2 & 245.6 & 0.608 & 0.773 & 3.14 & 2.87 & 0.752 \\

 + Progressive SAM  & 71.2 & 85.9 & 246.2 & 0.609 & 0.786 & 2.68 & 2.42 & 0.784 \\

 + Confidence-guided OPT & \textbf{69.4} & \textbf{83.6} & \textbf{218.4} & \textbf{0.709} & \textbf{0.816} & \textbf{2.53} & \textbf{2.34} & \textbf{0.789} \\

\bottomrule
\end{tabular}

\vspace{-0.3cm}
\caption{\textbf{Quantitative ablation studies on Hi4D.} We demonstrate the importance of the proposed progressive prompt for SAM and confidence-guided alternating optimization. Both key components effectively contribute to the final reconstruction quality (\cf \figref{fig:ablation_vis}).}
\label{tab:ablation}
\vspace{-0.5cm}
\end{table*}
}

\newcommand{\tableReconComp}{
\begin{table}[t]
\centering
\small

\resizebox{\columnwidth}{!}{%
\begin{tabular}{c l c c c c}
\toprule
\textbf{Dataset} & \textbf{Method} & $\textbf{V-IoU}$~$\uparrow$ & $\mathbf{C}-\ell_{2}$~$\downarrow$ & \textbf{P2S}~$\downarrow$ & \textbf{NC}~$\uparrow$  \\ 
\midrule

\multirow{3}*{Hi4D} & ECON \cite{xiu2023econ} & 0.787 & 3.72 & 3.59 & 0.746   \\ 
~ & V2A \cite{guo2023vid2avatar}  & 0.783 & 3.02 & 2.46 & 0.775   \\
~ & Ours & \textbf{0.816} & \textbf{2.53} & \textbf{2.34} & \textbf{0.789} \\
\midrule
\multirow{3}*{MMM} & ECON \cite{xiu2023econ}  & 0.760 & 4.17 & 3.71 & 0.705   \\
~ & V2A \cite{guo2023vid2avatar}  & 0.812 & 3.34 & 2.68 & 0.735   \\
~ & Ours & \textbf{0.826} & \textbf{2.89} & \textbf{2.40} & \textbf{0.757} \\
\bottomrule
\end{tabular}}

\vspace{-0.2cm}
\caption{\textbf{Quantitative reconstruction evaluation.} Our method consistently outperforms all baselines on both datasets and all evaluation metrics (\cf \figref{fig:recon}).}
\label{tab:recon}
\vspace{-0.4cm}
\end{table}
}

\newcommand{\tableProgressiveSAM}{
\begin{table}[t]
\centering
\small

\begin{tabular}{l c c c}
\toprule
\small
\textbf{Method} & \textbf{IoU}~$\uparrow$ & \textbf{Recall}~$\uparrow$ & \textbf{F1}~$\uparrow$ \\ 
\midrule

SCHP \cite{li2020self}  & 0.937 & 0.983 & 0.982   \\ 
\midrule
Ours (Init.) & 0.943 & 0.975 & 0.984  \\
Ours (Progressive) & \textbf{0.963} & \textbf{0.985} & \textbf{0.990}  \\
\bottomrule
\end{tabular}
\vspace{-0.1cm}
\caption{\textbf{Quantitative instance segmentation evaluation on Hi4D.} Our method achieves the best segmentation accuracy.}
\label{tab:seg}
\vspace{-0.5cm}
\end{table}
}

\newcommand{\tableNVS}{
\begin{table}[t]
\centering
\small

\begin{tabular}{l c c c}
\toprule
\textbf{Method} & \textbf{SSIM}~$\uparrow$  & \textbf{PSNR}~$\uparrow$ & \textbf{LPIPS}~$\downarrow$ \\ 
\midrule
Shuai et al. \cite{shuai2022multinb}  & 0.898 & 19.6 & 0.1099   \\ 
Ours & \textbf{0.915} & \textbf{20.7} & \textbf{0.0798}  \\
\bottomrule
\end{tabular}
\vspace{-0.2cm}
\caption{\textbf{Quantitative rendering evaluation on Hi4D.} Our method achieves better rendering quality (\cf \figref{fig:nvs}).}
\label{tab:novel}
\end{table}
}

\newcommand{\tableSupAblation}{
\begin{table*}[ht]
\centering
\small

\begin{tabular}{l|cccc|cccc}
\toprule
\multirow{2}*{Metrics} & \multicolumn{4}{c|}{\textbf{Pose Estimation}} & \multicolumn{4}{c}{\textbf{Human Reconstruction}} \\ 
\cline{2-9}
~ & \textbf{MPJPE}~$\downarrow$ & \textbf{MVE}~$\downarrow$ & \textbf{CD}~$\downarrow$ & \textbf{PCDR}~$\uparrow$ & \textbf{V-IoU}~$\uparrow$ & $\mathbf{C}-\ell_{2}$~$\downarrow$ & \textbf{P2S}~$\downarrow$ &\textbf{NC}~$\uparrow$ \\ 
\midrule
Initial pose & 75.3 & 90.8 & 235.6 & 0.566 & - & - & - & - \\
\midrule
Layer-wise volume rendering  & 70.8 & 85.1 & 248.6 & 0.605 & 0.746 & 3.85 & 3.69 & 0.719 \\

+ Progressive SAM & 70.7 & 85.0 & 247.7 & 0.609 & 0.793 & 2.72 & 2.44 & 0.777  \\

 + Confidence-guided OPT & \textbf{69.7} & \textbf{83.6} & \textbf{217.0} & \textbf{0.689} & \textbf{0.803} & \textbf{2.62} & \textbf{2.39} & \textbf{0.785} \\
\bottomrule
\end{tabular}
\caption{\textbf{Quantitative ablation studies on Hi4D based on tri-plane representation.} We demonstrate the importance of the proposed progressive prompt for SAM and confidence-guided alternating optimization. Both key components effectively contribute to the final reconstruction quality. This improvement generalize to different neural human representations.}
\label{tab:sup_ablation}
\vspace{-0.3cm}
\end{table*}
}

\newcommand{\tableReconCompRebuttal}{
\begin{table}[h]
\centering
\small
\vspace{-0.6cm}
\begin{tabular}{l c c c c}
\toprule
 \textbf{Method} & $\textbf{V-IoU}$~$\uparrow$ & $\mathbf{C}-\ell_{2}$~$\downarrow$ & \textbf{P2S}~$\downarrow$ & \textbf{NC}~$\uparrow$  \\ 
\midrule
 SLAHMR  & 0.813 & 3.71 & 3.52 & 0.739   \\
 Ours & \textbf{0.816} & \textbf{2.53} & \textbf{2.34} & \textbf{0.789} \\
\bottomrule
\end{tabular}
\vspace{-0.35cm}
\caption{\textbf{Reconstruction} evaluation with SLAHMR on Hi4D.}
\label{tab:reconR}
\vspace{-0.4cm}
\end{table}
}

\newcommand{\tableposeRebuttal}{
\begin{table}[h]
\centering
\small
\vspace{-0.38cm}

\begin{tabular}{l c c c c}
\toprule
\textbf{Method} & \textbf{MPJPE}~$\downarrow$ & \textbf{MVE}~$\downarrow$ & \textbf{CD}~$\downarrow$ & \textbf{PCDR}~$\uparrow$ \\ 
\midrule
SLAHMR  & 85.7 & 99.4 & 265.9 & 0.622  \\ 
Ours & \textbf{69.4} & \textbf{83.6} & \textbf{218.4} & \textbf{0.709}  \\
\bottomrule
\end{tabular}
\vspace{-0.35cm}
\caption{\textbf{Pose estimation} evaluation with SLAHMR on Hi4D.}
\label{tab:poseR}
\vspace{-0.5cm}
\end{table}
}

\newcommand{\tableProgressiveSAMRebuttal}{
\begin{table}[h]
\centering
\small
\vspace{-0.3cm}

\begin{tabular}{l c c c c}
\toprule
\small
\textbf{Method} & \textbf{IoU}~$\uparrow$ & 
\textbf{F1}~$\uparrow$ & \textbf{Recall}~$\uparrow$ & \textbf{Precision}~$\uparrow$ \\ 
\midrule
SCHP  & 0.905 & 0.972 & 0.973 & 0.972   \\ 
\midrule
Ours (Init.) & 0.912 & 0.975 & 0.992 & 0.959  \\
Ours (Progressive) & \textbf{0.951} & \textbf{0.987} & \textbf{0.994} & \textbf{0.979}  \\
\bottomrule
\end{tabular}
\vspace{-0.3cm}
\caption{\textbf{Instance segmentation} evaluation on ``contact'' Hi4D.}
\label{tab:segR}
\vspace{-0.5cm}
\end{table}
}

%% file: alg.tex
\newcommand{\algmulti}{
\begin{algorithm}
\caption{Optimization pipeline}
\label{alg:A}
\begin{algorithmic}
\STATE {$\mathcal{\theta} \gets $  Initial SMPL parameters} 
\STATE {$\mathcal{S} \gets $  Initial Neural Avatar}
\STATE {$\mathcal{M} \gets $  SAM($\mathcal{\theta}$, $\mathcal{S}$) (Initial SAM mask)}
\REPEAT

\STATE $\mathcal{\theta}, \mathcal{S}\gets $ Confidence-guided Opt($\mathcal{M}, \mathcal{\theta}, \mathcal{S}$) (Sec.\ref{sec:delay})

\STATE $\mathcal{M} \gets $ SAM($\mathcal{\theta}, \mathcal{S}$) (Sec.\ref{sec:sam_prompt})
\UNTIL{max loop reached}

\end{algorithmic}
\end{algorithm}
}

%% file: sec/00_abstract.tex
\begin{abstract}
\setlength{\topsep}{0pt}
\setlength{\parskip}{.5ex}
\renewcommand{\floatsep}{1ex}
\renewcommand{\textfloatsep}{1ex}
\renewcommand{\dblfloatsep}{1ex}
\renewcommand{\dbltextfloatsep}{1ex}

We present \methodname, a novel framework to reconstruct multiple people in 3D from monocular in-the-wild videos. Reconstructing multiple individuals moving and interacting naturally from monocular in-the-wild videos poses a challenging task. Addressing it necessitates precise pixel-level disentanglement of individuals without any prior knowledge about the subjects. Moreover, it requires recovering intricate and complete 3D human shapes from short video sequences, intensifying the level of difficulty. To tackle these challenges, we first define a layered neural representation for the entire scene, composited by individual human and background models. We learn the layered neural representation from videos via our layer-wise differentiable volume rendering. This learning process is further enhanced by our hybrid instance segmentation approach which combines the self-supervised 3D segmentation and the promptable 2D segmentation module, yielding reliable instance segmentation supervision even under close human interaction. A confidence-guided optimization formulation is introduced to optimize the human poses and shape/appearance alternately. We incorporate effective objectives to refine human poses via photometric information and impose physically plausible constraints on human dynamics, leading to temporally consistent 3D reconstructions with high fidelity. The evaluation of our method shows the superiority over prior art on publicly available datasets and in-the-wild videos.
\end{abstract}

%% file: sec/01_introduction.tex
\section{Introduction}

Despite rapid progress in estimating the 3D shape from monocular videos of a single performer \cite{guo2023vid2avatar,  saito2020pifuhd, xiu2022icon, jiang2023instantavatar}, the analysis and reconstruction of several closely interacting people is still limited.
This imbalance is unsatisfying, as group activities make up a significant portion of our lives. 
Although systems for multi-person reconstruction have been previously investigated, most require multi-view setups that constrain the capture area to a fixed volume and demand specialized equipment and expertise to operate \cite{shuai2022multinb, zhang2021editable, yin2023hi4d}.
Being able to reconstruct multiple people in detailed 3D geometry and appearance from monocular videos -- which is also amenable to novice users -- would facilitate many downstream tasks in AR/VR, such as the telepresence of groups of people, or the ``replay'' of social activities in 4D.
Accomplishing this is fundamentally a challenging task since it requires accurate pixel-level disentanglement of individuals without a priori known geometries of the subjects.
To make matters worse, the task is further complicated by depth ambiguities, complex human dynamics, and severe human-human occlusions -- all of which have to be resolved from a single, short video clip.

In this paper, we introduce a novel method, called \methodname, that provides a solution to this task: It takes a single video as input and outputs complete, high-quality, and separated 3D human geometry and appearance for individuals appearing in the scene (see \figref{fig:teaser}).
By embracing the promising paradigm of neural implicit functions for 3D representations, remarkable progress has recently been achieved in modelling detailed human geometry and appearance.
While some methods require full supervision on 3D human scans that can be prohibitively expensive to acquire \cite{saito2020pifuhd, xiu2023econ}, others only rely on readily available monocular videos \cite{guo2023vid2avatar, Feng2022scarf, jiang2022selfrecon, weng_humannerf_2022_cvpr, jiang2022neuman} to fit articulated neural implicit fields with temporally consistent results.
However, all of these methods are designed for only a single actor and thus neglect the complexities of the reconstruction task that arise from severe human-human occlusions.
When applied to footage where multiple people engage in natural interactions, the aforementioned methods often result in corrupted or incomplete human reconstructions.

Our method builds on the promise of neural implicit fields, and presents a solution that overcomes the limitations of prior work.
In doing so, several challenges must be addressed.
First, 2D and 3D points of each subject must be precisely associated.
Second, complete human models must be extracted and maintained from only a single video.
Third, and most critically, the problem of person association and avatar creation are both significantly exacerbated by strong occlusions and imperfect human pose estimates.

To tackle these challenges, our approach is grounded in the following core concepts:
\begin{inparaenum}[i)]
    \item We design a unified, temporally consistent representation of human shape and texture in canonical space that is applicable to all individuals. This design facilitates the integration of partial observations from the video sequence into a temporally coherent space, inherently maintaining a complete human body.
    \item We establish a layered neural representation for the entire scene, wherein we parameterize the humans and the background as individual neural fields. The composition of these fields leads to a layered, interwoven representation that covers the entire space and which can be learned from a monocular RGB video via a proposed layer-wise differentiable volume rendering.
    \item  We introduce a hybrid instance segmentation approach that leverages the advantages of self-supervised scene decomposition in 3D and a learning-based promptable 2D segmentation module. This combination results in a robust and accurate separation of individuals.
    \item To deal with imperfect pose estimates that might corrupt updates to the avatar model, we present a confidence-guided optimization formulation that alternately optimizes human poses and shape/appearance based on per-frame confidence measures. This way we incorporate effective objectives that refine the human poses through photometric information and impose physically plausible constraints on human dynamics.
\end{inparaenum}

In our experiments, we demonstrate that our framework leads to robust human instance segmentation and plausible pose estimates, achieving high-quality 3D reconstructions of multiple people even under extremely challenging visual complexities like severe occlusions (\cf \figref{fig:method}). We meticulously ablate our method, which uncovers its essential components.
Furthermore, we conduct comparisons with existing approaches in human reconstruction, novel view synthesis, human instance segmentation, and pose estimation tasks, showing that our method outperforms prior art across various settings.
In summary, our contributions are:
\begin{compactitem}

 \item a novel framework, \methodname, to reconstruct multiple detailed 3D human models solely from in-the-wild monocular videos; and
 \item a robust instance segmentation approach that achieves a clean separation between people even under close interaction; and
 \item a confidence-guided optimization formulation that leads to temporally and spatially coherent 3D reconstructions of people with high fidelity.

\end{compactitem}

%% file: sec/02_related_work.tex
\vspace{-0.1cm}
\section{Related Work}

\paragraph{Monocular Single-Person Reconstruction}

Reconstructing an individual from monocular observations has emerged as a widely explored research challenge. Template-based approaches involve the tracking of a human template using 2D observations \cite{deepcap}. The assumption of a personalized template is unsuitable for more practical use cases. Follow-up works endeavor to remove this dependency by adding the vertex offsets on top of SMPL \cite{alldieck2018video, guo2021human}. Nevertheless, the explicit mesh representation is constrained by a fixed resolution and topology, incapable of representing fine-grained details. Learning-based methods that learn to regress 3D human shape from images have shown compelling results \cite{saito2019pifu, huang2020arch, saito2020pifuhd, zheng2021pamir, xiu2022icon, alldieck2022phorhum, xiu2023econ}. A major limitation of these methods is the necessity of high-quality 3D data for supervision and they fail to produce space-time coherent reconstructions over frames. Recent works employ neural rendering to train neural fields based on videos to obtain articulated human model \cite{su2021anerf, peng2021neural, jiang2022selfrecon, weng_humannerf_2022_cvpr, jiang2022neuman, Feng2022scarf, guo2023vid2avatar, jiang2023instantavatar, Feng2023DELTA}. Among these, Vid2Avatar \cite{guo2023vid2avatar} achieves compelling 3D reconstruction for a single subject but is not directly applicable to scenes with crowded people. Actually, none of the aforementioned methods can be directly deployed in more complicated multi-person scenarios. In contrast, we propose a novel framework that can faithfully reconstruct multiple people in the scene from a monocular video.
\vspace{-0.4cm}
\paragraph{Monocular Multi-Person Reconstruction}
In contrast to the notable advancements in reconstructing the clothed human for an individual, limited emphasis has been placed on multi-person scenarios, which are evidently more applicable to our daily experiences. 
Most existing monocular works can only estimate the coarse body shapes of multiple people from monocular observations \cite{jiang2020coherent, fieraru2020chi, kocabas2021pare, sun2022bev, TRACE, khirodkar2022occluded, Cheng_2023, huang2023reconstructing, li2023jotr, ye2023slahmr}. 
Mustafa et al. \cite{mustafa2021multi} extend prior implicit methods to multiple people and recover spatially coherent 3D human shapes from an RGB image but mainly deal with cases where people are well-spaced and do not interact naturally in close range. 
Recently, more researchers have shifted the focus to multi-person scenarios \cite{zhang2021stnerf, zheng2021deepmulticap, shuai2022multinb, yin2023hi4d}.
Even though these works achieve compelling instance-level human reconstructions, they require expensive multi-view imaging systems. 
Concurrently, Cha et al. \cite{cha20243d} propose to reconstruct multiple people from a single image. Such image-based methods usually fail to produce space-time coherent reconstructions over frames.
Overall, monocular multi-person reconstruction is still an extremely under-explored problem. We propose MultiPly to take a significant stride towards addressing this formidable task.
\vspace{-0.6cm}
\paragraph{Human Instance Segmentation}
Most works solve human or general object segmentation at the image level (i.e. 2D) \cite{he2017maskrcnn, kirillov2020pointrend, li2020self, wu2019detectron2}. 
They are trained on images with human annotations to directly regress the segmentation masks during inference. 
More recently, a promptable segmentation model SAM has been developed to support flexible prompting along with input images \cite{Kirillov_2023_ICCV}.
However, SAM is a semantic segmentation method, which does not inherently support instance-level segmentation for humans. Therefore, meticulous prompts need to be designed for human instance segmentation tasks.
Besides, these approaches are not able to produce sharp boundaries between individuals, especially for closely interacting people. 
More importantly, they do not always predict temporally coherent segmentation masks, as they focus on image-level segmentation only and incorporate no 3D knowledge.
In this work, we optimize the instance segmentation masks on the fly by leveraging the promptability of SAM \cite{Kirillov_2023_ICCV} and the self-supervised decomposition in 3D \cite{shuai2022multinb, guo2023vid2avatar}.

%% file: sec/03_method.tex
\vspace{-0.2cm}
\section{Method}
\figurePipeline
We present \methodname, a novel framework for detailed geometry and appearance reconstruction of multiple people from in-the-wild monocular videos. The overview of our method is schematically illustrated in~\figref{fig:method}. Reconstructing multiple people in 3D from a short video without a priori known geometries is a challenging task due to complex human articulation, and strong occlusions. To tackle these challenges, we first define a unified, temporally consistent representation for humans and a layered neural representation for the entire scene (\secref{sec:representation}). The layered neural representation is then learned from images by performing our layer-wise differentiable volume rendering (\secref{sec:volume}). Given the self-supervised instance segmentation via occlusion-aware volume rendering, we further enhance the instance segmentation supervision by leveraging our evolving human surfaces in deformed space as progressively updated prompts for SAM  which builds closed-loop refinement of instance segmentation in both 2D and 3D (\secref{sec:sam_prompt}). Finally, we formulate a confidence-guided optimization to alternately optimize human pose and shape/appearance (\secref{sec:multi}). We incorporate photometric information, robust instance segmentation supervision, and the inter-person objectives for pose refinement to achieve temporally and spatially coherent 3D reconstructions of people in high quality (\secref{sec:obj}).

\subsection{Layered Neural Representation}
\label{sec:representation}
\paragraph{Neural Avatars.}
For each human in the scene, we represent the 3D shape as an implicit signed-distance field (SDF) and the appearance as a texture field in canonical space, covering the entire space. When multiple people are present in the scene, it leads to a layered representation where the contributing SDFs are potentially interwoven. More specifically, we model the geometry and appearance of each person $p$ in canonical space by a neural network $f^{p}$, which predicts the signed distance value $s^{p}$ and the radiance value $\boldsymbol{c}^{p}$ at the query point $\boldsymbol{x}_c^p$ in this space:
\begin{align}
\boldsymbol{c}^{p}, s^{p} & = f^{p}(\boldsymbol{x}_c^p, \boldsymbol{\theta}^{p}),
\end{align}
where $\boldsymbol{\theta}^{p}$ denotes the person's pose parameters, which we concatenate to $\boldsymbol{x}_c^p$ to model pose-dependent surface deformations. For simplicity, we use $f_c^p(\cdot)$ and $f_s^p(\cdot)$ to query $\boldsymbol{c}^{p}$ and $s^{p}$ from the network outputs.
\vspace{-0.2cm}
\paragraph{Deformation Module.} We follow a standard skeletal deformation based on SMPL \cite{loper2015smpl} to find correspondences in canonical and deformed space.
A canonical point $\boldsymbol{x}_c^p$ is mapped to the deformed point $\boldsymbol{x}_d^p$ via linear-blend skinning (LBS) based on SMPL transformation: $\boldsymbol{x}_d^p = T_{\text{smpl}}(\boldsymbol{x}_c^p, \boldsymbol{\theta}^{p})$.
Here, $T_{\text{smpl}}(\cdot)$ denotes the SMPL-based transformation derived from the body pose $\boldsymbol{\theta}^{p}$, which corresponds to LBS and is described in more detail in the \suppmat.
Inversely, the canonical correspondence $\boldsymbol{x}_c^p$ for point $\boldsymbol{x}_d^p$ in deformed space is defined as $\boldsymbol{x}_c^p = T_{\text{smpl}}^{-1}(\boldsymbol{x}_d^p, \boldsymbol{\theta}^{p})$. To invert LBS we use the SMPL skinning weight of the vertex closest to $\boldsymbol{x}_d^p$.

\subsection{Layer-Wise Volume Rendering}
\label{sec:volume}
We seek to reconstruct all human subjects in the scene. This requires different treatment compared to vanilla differentiable volume rendering that only works on a single static scene \cite{mildenhall2020nerf}. In contrast, on the basis of our layered neural avatar representation (\secref{sec:representation}), we introduce layer-wise volume rendering to handle dynamic scenes with multiple subjects and inter-occlusions. This is achieved by combining surface-based volume rendering \cite{yariv2021volume} with the re-assembly of multiple human neural layers \cite{zhang2021stnerf}. It is essential to note that the layer-wise volume rendering is inherently occlusion-aware.
\paragraph{Volume Rendering for Human Layers.} 
For each sampled camera ray $\boldsymbol{r}$, we sample the points in the observation space along the ray based on the intersection between the oriented bounding box of the deformed SMPL model and the camera ray. Specifically, we sample $N$ points $\{\boldsymbol{x}_{d,1}^p,...,\boldsymbol{x}_{d,N}^p\}$ inside the $p$-th intersected bounding box based on the two-stage sampling strategy proposed in \cite{yariv2021volume}. Then, the occupancy $o^{p}_{i}$ for the $p$-th person and the $i$-th sampled point is calculated as follows:
\begin{align}
o^{p}_{i} & = \left(1-\exp \left(-\sigma_i^p \Delta\boldsymbol{x}_i\right)\right), \nonumber \\
\sigma_i^p & = \sigma \left (f_s^p \left (  T_{\text{smpl}}^{-1} \left (\boldsymbol{x}_{d,i}^p , \boldsymbol{\theta}^p \right), \boldsymbol{\theta}^p \right  ) \right ),
\end{align}
where $\Delta\boldsymbol{x}_i$ is the distance between two adjacent sample points, and $\sigma(\cdot)$ is the scaled Laplace distribution’s Cumulative Distribution Function (CDF) defined in \cite{yariv2021volume} to convert the signed distance $s_i^p$ to volume density $\sigma_i^p$. Then we accumulate the radiance by performing numerical quadrature among the layered density field for multiple persons to obtain the color value:
\begin{equation}
\label{eq:color}
\hat{C}^H=\sum_{i=1}^{N} \sum_{p=1}^{P} \left[ o_i^p  \boldsymbol{c}_i^p \prod_{q=1}^{P} \prod_{j \in \mathcal{Z}_{i}^{q,p}}\left(1-o_j^q \right) \right],
\end{equation}
where $P$ is the total number of subjects, $\hat{C}^H$ is the rendered color of humans, and $\mathcal{Z}_i^{q,p}$ contains all indices of points (belonging to person $q$) whose depth value is lower than the depth value of the $i$-th point of person $p$, \ie $\mathcal{Z}_{i}^{q,p} = \{ j \in [1, N] \mid z(\boldsymbol{x}_{d,j}^q)<z(\boldsymbol{x}_{d,i}^p)\}$, where $z(\cdot)$ denotes the distance between the sampled point and the camera origin along the z-axis.

\paragraph{Scene Composition.}
We model the background in the same formulation as NeRF++ \cite{kaizhang2020}, denoted as $f^b$. We thus obtain a color value $\hat{C}^B$ representing the background's color, which is composited with $\hat{C}^H$ via self-supervised decomposition following \cite{guo2023vid2avatar} to compute the final pixel color value $\hat{C}$. More details are shown in the \suppmat.

\subsection{Progressive Prompt for SAM}
\label{sec:sam_prompt}
Learning to disentangle and reconstruct multiple subjects by simply relying on the automatic separation through layer-wise volume rendering is still a severely ill-posed problem. This is due to dynamically changing lighting effects (e.g., shadows) and potentially severe human-human occlusions and close contact. To this end, we propose to leverage the promptable segmentation model SAM \cite{Kirillov_2023_ICCV} and design a progressive prompting strategy based on the evolving human models to provide robust instance segmentation supervision. In this section we describe how we design the prompt to get an updated SAM mask which we later use in the optimization (\secref{sec:multi} and \secref{sec:obj}).

We define the $p$-th human shape to be the zero-level set of the signed distance function $f_s^p$ and apply the \textit{Multiresolution IsoSurface Extraction} (MISE) \cite{mescheder2019occupancy} to extract the mesh in canonical space, denoted as $\mathcal{S}_c^p=\left \langle \mathcal{V}_c^p, \mathcal{F}^p  \right \rangle = \text{MISE}(f_s^p, \boldsymbol{\theta}^p)$. Here, $\mathcal{V}_c^p$ represents the extracted vertex set in canonical space, and $\mathcal{F}^p$ denotes the corresponding faces.
Then, the deformed vertex set in the observation space is:
\begin{equation}
\label{eq:dv}
\mathcal{V}_d^p=\left\{T_{\text{smpl}}(\boldsymbol{v}_c^p, \boldsymbol{\theta}^{p}) \mid \boldsymbol{v}_c^p \in \mathcal{V}_c^p \right\}.
\end{equation}
Similarly, the deformed mesh for the $p$-th person in the observation space is defined as $\mathcal{S}_d^p=\left \langle \mathcal{V}_d^p, \mathcal{F}^p  \right \rangle $. Thus, we can obtain an instance mask $\mathcal{M}_{\text{mesh}}^p$ by differentiably rendering the deformed mesh. To improve efficiency, we opt for a differentiable rasterizer $R$ rather than volume rendering:
\begin{equation}
\mathcal{M}_{\text{mesh}}^p=R (\mathcal{S}_d^p).
\end{equation}
For the sake of clarity, we define $\mathcal{M}=1$ to represent the inside and $\mathcal{M}=0$ to indicate either the outside or occlusion by other meshes.
The instance mask of deformed meshes $\mathcal{M}_{\text{mesh}}^p$ serves as one of the prompts for SAM refinement.
We further provide points as input prompts.
We begin by projecting the 3D keypoint candidates onto the image to obtain 2D keypoints $\mathcal{K}_{2d}^p=\left\{\Pi\left ( \mathcal{J} (\boldsymbol{\theta}^{p}, \boldsymbol{\beta}^{p})\right )\right\}$, where $\mathcal{J}(\boldsymbol{\theta}^{p}, \boldsymbol{\beta}^{p})$ is the 3D SMPL keypoints given the pose $\boldsymbol{\theta}^{p}$ and body shape $\boldsymbol{\beta}^p$ parameters for subject $p$, and $\Pi$ is the camera projection function.
The point prompts for the $p$-th subject are then defined by:  
\begin{equation}
\begin{aligned}
 & \mathcal{P}_{+}^p=\left\{ \boldsymbol{k} \in \mathcal{K}_{2d}^p \mid  \mathcal{M}_{\text{mesh}}^p(\boldsymbol{k})=1 \right\}, \\
 & \mathcal{P}_{-}^p=\{ \boldsymbol{k} \in  {\textstyle \bigcup_{q \ne p}} \mathcal{K}_{2d}^q \mid  \mathcal{M}_{\text{mesh}}^p(\boldsymbol{k})=0 \}.
\end{aligned}
\end{equation}
In other words, the positive point prompts $\mathcal{P}_{+}^p$ include the 2D keypoints that are inside of the instance mask obtained from the deformed mesh and are outside of the instance masks of all other meshes. The negative point prompts are the union of all 2D keypoints of all other subjects that are outside of the projected mesh mask $\mathcal{M}_{\text{mesh}}^p$.
The SAM instance mask $\mathcal{M}_\text{sam}^p$ is finally updated based on the combination of the mask and point prompts:
\begin{equation}
\mathcal{M}_{\text{sam}}^p = \text{SAM}(\mathcal{M}_{\text{mesh}}^p, \mathcal{P}_{+}^p, \mathcal{P}_{-}^p),
\end{equation}
Note that $\mathcal{M}_\text{sam}^p$ are progressively updated during training.

\subsection{Confidence-Guided Alternating Optimization}
\label{sec:multi}
Human-human occlusions often lead to inaccurate pose and wrong depth order estimation. A na\"ive joint optimization for both the pose and shape parameters may end up with a suboptimal solution. To mitigate this, we introduce a confidence-guided optimization strategy to alternately optimize the human poses and shapes. 

\label{sec:delay}
To avoid damaging shape updates that are due to wrong poses, we only optimize pose parameters for unreliable frames and jointly optimize pose \emph{and} shape parameters for reliable frames.
We treat the IoU between the projected mesh mask $\mathcal{M}_{\text{mesh}}^{p,i}$ and the refined SAM mask $\mathcal{M}_{\text{sam}}^{p,i}$ as our confidence measure for the $p$-th subject in frame $i$. We define reliable frames $\mathcal{I}_r$ to be those frames with reliable poses based on the average IoU over all subjects:
\begin{equation}
\mathcal{I}_r = \left \{ \boldsymbol{I}_i \in \mathcal{I} \mid \frac{1}{P} \sum_{p=1}^{P}\text{IoU}(\mathcal{M}_{\text{mesh}}^{p,i}, \mathcal{M}_{\text{sam}}^{p,i}) \ge \alpha \right \}, 
\end{equation}
where $\mathcal{I}$ are all frames, and $\mathcal{I} \setminus \mathcal{I}_r$ are unreliable frames. $\alpha$ is a confidence threshold which is set to be the median of all IoU values over the entire sequence. It's important to note that the confidence threshold $\alpha$ is dynamically updated during training and eventually inaccurate pose estimates are corrected and all frames will be used for joint optimization. 
\tableAblation

\subsection{Objectives}
\label{sec:obj}

\noindent\textbf{Reconstruction Loss.} We calculate the $L_{1}$-distance between the rendered color $\hat{C}(\mathbf{r})$ and the image pixel's RGB value $C(\mathbf{r})$ over all sampled rays $\mathcal{R}$:
\begin{equation}
L_{\text{rgb}} = \frac{1}{|\mathcal{R}|} \sum_{\boldsymbol{r} \in \mathcal{R}} | C(\boldsymbol{r}) - \hat{C}(\boldsymbol{r})|.
\end{equation} 

\noindent\textbf{Instance Mask Loss.} We first modify Eq.~\ref{eq:color} to differentiably render the opacity $\hat{O}^p(\boldsymbol{r})$ per person per pixel:
\begin{equation}
\hat{O}^p(\boldsymbol{r})=\sum_{i=1}^{N} \left[ o_i^p \prod_{q=1}^{P} \prod_{j \in \mathcal{Z}_{i}^{q,p}}\left(1-o_j^q \right) \right].
\end{equation}
Then the instance mask loss is calculated between the refined instance mask and the rendered pixel-wise opacity:
\begin{equation}
L_{\text{mask}} = \frac{1}{|\mathcal{R}|} \sum_{\boldsymbol{r} \in \mathcal{R}} \sum_{p=1}^{P} | \mathcal{M}_{\text{sam}}^{p}(\boldsymbol{r}) - \hat{O}^{p}(\boldsymbol{r})|.
\end{equation}

\noindent\textbf{Eikonal Loss.} Following \cite{gropp2020implicit}, we sample points in the canonical space for each subject and enforce the Eikonal constraint to ensure $f_s^p$ is a valid SDF:
\begin{equation}
L_{e}=\sum_{p=1}^{P} \mathbb{E}_{\boldsymbol{x}_c}\left(\left\|\nabla f_s^p \left(\boldsymbol{x}_c^p, \boldsymbol{\theta}^p \right)\right\|-1\right)^2.
\end{equation}
We further introduce two inter-person objectives for pose refinement to ensure spatially coherent and physically plausible reconstructions. We apply these constraints explicitly on the deformed mesh to refine human poses while the deformed meshes $\mathcal{S}_d^p$ are updated on the fly during training. 
To be specific, the following two additional losses are used periodically during training to optimize pose only:

\noindent\textbf{Depth Order Loss.} The effect of wrong depth order on the reconstruction quality can be severe, resulting in reversed geometry and texture. Similar to \cite{jiang2020coherent}, we apply a depth order loss as follows:
\begin{equation}
L_{\text{depth}} = \sum_{(\boldsymbol{u}, p, q) \in \mathcal{D}} \log (1+\exp (D_p(\boldsymbol{u})-D_q(\boldsymbol{u}))),
\end{equation}
where $\mathcal{D} = \left \{ (\boldsymbol{u}, p, q) \mid p \ne q, \mathcal{M}_{\text{sam}}^p(\boldsymbol{u})=\mathcal{M}_{\text{mesh}}^q(\boldsymbol{u})=1 \right  \}$ represents the set of pixels $\boldsymbol{u}$ where we have depth ordering mistakes between the $p$-th and $q$-th persons. $D_p(\boldsymbol{k})$ denotes the depth of the $p$-th mesh for pixel $\boldsymbol{u}$.

\noindent\textbf{Interpenetration Loss.} We shoot a ray for the sampled vertex in $\mathcal{V}_d^p$ in Eq. \ref{eq:dv} to check the number of intersection with other meshes. Then, we use the parity of the number of intersections to determine whether that point is inside other meshes. Following \cite{NEURIPS2021_a1a2c3fe}, the interpenetration loss is calculated as follows: 
\begin{equation}
L_{\text{inter}} = \sum_{p=1}^{P} \sum_{q=1, q\ne p}^{P} \left \| \mathcal{V}_{\text{in}}^{p,q} - NN(\mathcal{V}_{\text{in}}^{p,q}, S_d^q)  \right \|_2 ,
\end{equation}
where $\mathcal{V}_{\text{in}}^{p,q}$ denotes the $p$-th person's vertex which is inside the $q$-th person's mesh, and $NN(\mathcal{V}, \mathcal{S})$ finds the nearest vertex in $\mathcal{S}$ for each point in $\mathcal{V}$. Different from \cite{NEURIPS2021_a1a2c3fe,jiang2020coherent} where they deploy the depth order and interpenetration loss on the naked parametric human model, we apply those two losses on our learned pixel-wise aligned clothed human meshes, leading to a more fine-grained optimization.

See \suppmat for more details about the final loss.

%% file: sec/04_experiment.tex
\figureAblation
\section{Experiments}
We first introduce the datasets and metrics used for evaluation. Next, ablation studies are conducted to demonstrate the effectiveness of our design choices. We then compare our proposed method with state-of-the-art approaches in four tasks, including human reconstruction, novel view synthesis, human instance segmentation, and pose estimation. 
\subsection{Datasets and Metrics}
\label{sec:dataset}
\noindent\textbf{Hi4D \cite{yin2023hi4d}.} This dataset contains challenging human interactions between pairs of people with ground truth meshes, human poses, and instance segmentation masks.  We use Hi4D to evaluate our approach to all tasks. 

\noindent\textbf{Monocular Multi-huMan (MMM).}
Since the Hi4D dataset only contains two-person interactions with the static camera in the stage. In order to evaluate the generalization of our method, we collect a dataset called \textit{Monocular Multi-huMan (MMM)} by using a hand-held smartphone, which contains six sequences with two to four persons in each sequence. Half of the sequences are captured in the stage with ground truth annotations for quantitative evaluation and the others are captured in the wild for qualitative evaluation. 

\noindent\textbf{Metrics.}
We consider the following metrics for human mesh reconstruction evaluation: volumetric IoU (V-IoU), Chamfer distance ($\mathbf{C}-\ell_{2}$) [cm], point-to-surface distance (P2S) [cm], and normal consistency (NC). Rendering quality is measured via three metrics: PSNR, SSIM, and LPIPS. We assess human pose estimation using four metrics: MPJPE [mm], MVE [mm], Contact Distance (CD) [mm], and Percentage of Correct Depth Relations (PCDR) with a threshold of 0.15m. Lastly, we report IoU, Recall, and F1 score for human instance segmentation.
\subsection{Ablation study}
As depicted in \tabref{tab:ablation}, ablation studies are conducted to demonstrate the effectiveness of the proposed progressive SAM prompt and confidence-guided optimization strategy. Both pose estimation and human reconstruction tasks are evaluated here. The initial human poses are obtained from TRACE \cite{TRACE} and ViTPose \cite{xu2022vitpose}, more details are provided in the \suppmat.
We initiate ablation studies based on our layer-wise volume rendering by naively optimizing the human pose and shape/appearance jointly without using SAM.
\vspace{-0.7cm}
\paragraph{Human Reconstruction.} Applying instance mask supervision based on progressively refined SAM outputs significantly improves the output quality and drastically reduces the reconstruction error (Chamfer distance) as quantitatively shown in \tabref{tab:ablation}. This is also confirmed by the qualitative results within the \textcolor[RGB]{255,125,0}{orange} bounding boxes highlighted in \figref{fig:ablation_vis}, where layer-wise volume rendering purely relying on self-supervised segmentation fails to separate the dynamic shadows from the human, leading to noisy reconstructions. The reconstruction quality is further improved with the proposed confidence-guided optimization, as quantitatively indicated by the last row of \tabref{tab:ablation}. The \textcolor[RGB]{255,0,0}{red} bounding boxes in \figref{fig:ablation_vis} serve as visual evidence for the importance of our confidence-guided optimization. The presence of incomplete human bodies, such as the broken leg, disappeared back, and shrunken neck, is attributed to the incorrect depth order and pose estimation error. By temporarily freezing the implicit network for frames with unreliable poses, we circumvent such detrimental shape updates, leading to a complete human reconstruction.
\vspace{-0.3cm}
\paragraph{Pose Estimation.} We observe that this performance aligns with geometric reconstruction. Compared with the initial pose, we achieve not only more accurate individual poses (MPJPE and MVE) but also a better spatial arrangement between people, reflected on CD and PCDR.
\label{sec:ablation}

\vspace{-0.1cm}
\subsection{Reconstruction Comparisons}
\label{sec:reconcomp}
To the best of our knowledge, there are few video-based reconstruction methods designed for clothed multiple people. Hence, we adapt two state-of-the-art reconstruction approaches to our setting for comparison. ECON \cite{xiu2023econ} is an image-based 3D human reconstruction method capable of handling multi-person scenarios. While evaluating ECON, we discard frames with incorrect bounding-box detections for a fair comparison. Vid2Avatar (V2A) \cite{guo2023vid2avatar} is a video-based human reconstruction method designed for a single person. We extend V2A to multi-person scenarios by training a distinct model for each subject individually. 
Our method outperforms \cite{xiu2023econ, guo2023vid2avatar} by a substantial margin on both datasets and all metrics (\cf \tabref{tab:recon}). This disparity becomes more visible in qualitative comparisons shown in \figref{fig:recon}. When people closely interact, both ECON and V2A fail to recover complete human bodies but only output corrupted reconstructions (\eg, missing legs/heads). Furthermore, they struggle with the initial depth order/pose error and produce spatially incorrect reconstructions in 3D. V2A tends to model environmental dynamic effects (\eg, shadows) as the human body, resulting in noisy reconstructions. These artifacts are highlighted within the colored bounding boxes of \figref{fig:recon}. In contrast, our method generates complete human shapes with sharp boundaries and spatially coherent 3D reconstructions. We attribute this superiority to our proposed representation design and learning schemes.
\figureReconComp
\tableReconComp
\subsection{Novel View Synthesis Comparisons}

To the best of our knowledge, there are few novel view synthesis approaches particularly designed for clothed multiple people from monocular video. Hence, we adapt Shuai et al. \cite{shuai2022multinb}, which is a state-of-the-art multi-person novel view synthesis approach from multi-view videos, to the monocular setting for a fair comparison.
We share the same human pose initialization for training. Then, we use the ground truth human poses and camera parameters to render the novel view. As shown in \tabref{tab:novel} our method outperforms \cite{shuai2022multinb} on all metrics. \figref{fig:nvs} shows that the rendered image from \cite{shuai2022multinb} is more blurry and has noisy artifacts compared to ours. The reasons are twofold: 1) it lacks a reliable pose correction mechanism, leading to large inconsistency between human pose and image information during training, and 2) weekly-supervised decomposition cannot ensure robust instance segmentation under close human interaction. Our framework generates more plausible renderings with clearly sharp boundaries.
\tableNVS
\figureNVS
\subsection{Instance Segmentation Comparisons}
We compare our instance segmentation result from SAM after convergence with pretrained human instance segmentation network SCHP \cite{li2020self}. \tabref{tab:seg} reveals that our initial SAM outputs achieve comparable results with SCHP. However, the initial SAM masks are unsatisfactory due to the noisy prompt from the inaccurate SMPL estimation, leading to an incomplete and implausible reconstruction result (\eg, the missing body part and self-interpenetration in the \textcolor[RGB]{255,0,0}{red} bounding boxes) as shown in \figref{fig:seg}. Finally, our progressive prompting strategy based on our evolving human models helps to achieve temporally consistent and complete segmentation masks and high-quality reconstructions, surpassing the baseline methods.

\tableProgressiveSAM

\figureProgressiveSAM

\subsection{Pose Estimation Comparisons}
We conduct a comparison of our method with state-of-the-art bottom-up (TRACE \cite{TRACE}) and top-down (CLIFF \cite{li2022cliff}) multi-person pose estimation approaches. To adapt CLIFF for close human interaction, we employ ByteTrack \cite{zhang2022bytetrack} and linear interpolation to estimate missing persons caused by detector errors. Our approach consistently outperforms other baseline methods on all metrics as shown in \tabref{tab:pose}. Specifically, our method shows its superiority in pose estimation accuracy of individuals (MPJPE and MVE) and more reasonable spatial arrangement between pairs of people (CD and PCDR). This is also confirmed by qualitative results. Please refer to the \suppmat.

\tablemask

%% file: sec/05_conclusion.tex
\vspace{-0cm}
\section{Conclusion}
In this paper, we present \methodname, which for the first time produces temporally and spatially coherent 3D reconstructions of multiple people with high fidelity from monocular in-the-wild videos. We utilize carefully designed layered neural representation and dynamically refined instance segmentation supervision. We further introduce a confidence-guided optimization to learn human neural layers via layer-wise volume rendering. Our method is able to reconstruct multiple high-quality 3D human models in challenging scenarios involving close human interactions and strong inter-person occlusions.

\noindent\textbf{Limitations:} The complexity of our model increases linearly with the number of involved persons, making it inefficient for crowds. Our method does not explicitly model hands and we see the integration of an expressive human model \cite{SMPL-X:2019} as a future direction. We discuss more limitations and potential negative societal impact in \suppmat.

\noindent\textbf{Acknowledgement:} This work was partially supported by the Swiss SERI Consolidation Grant ``AI-PERCEIVE".